\documentclass{article}

\usepackage[main, final]{neurips_2026}

\usepackage[utf8]{inputenc} 
\usepackage[T1]{fontenc}    
\usepackage{hyperref}       
\usepackage{url}            
\usepackage{booktabs}       
\usepackage{amsfonts}       
\usepackage{nicefrac}       
\usepackage{microtype}      
\usepackage{multirow}
\usepackage{array}
\usepackage{graphicx} 
\usepackage{amsmath}  
\usepackage{capt-of}
\usepackage{wrapfig}
\usepackage{caption}
\usepackage{subcaption}  
\usepackage{adjustbox}
\usepackage{xcolor}         
\usepackage{soul}
\usepackage{amssymb}
\usepackage{bbm}

\title{EgoTac: In-the-wild Tactile Prediction from Egocentric Vision}

\author{
  Wenkang Zhang$^{1,2}$ \quad Chengbo Yuan$^{3}$\quad Zicheng Zhang$^{4}$\quad Zhengxue Cheng$^{2}$\quad Yang Gao$^{1,3}$\thanks{Corresponding author} \\
  $^{1}$Shanghai Qi Zhi Institute  $^{2}$Shanghai Jiao Tong University \\
  $^{3}$Tsinghua University  $^{4}$Fudan University \\
  {\tt\small conquer.wkzhang@sjtu.edu.cn, gaoyangiiis@mail.tsinghua.edu.cn}
}

\begin{document}

\maketitle

\begin{abstract}
Touch is fundamental to dexterous manipulation, yet most egocentric human data increasingly used for robot learning lacks tactile information. 
Directly collecting large-scale tactile data is challenging due to sensor limitations, while human video data is abundant, contact-rich, and easily scalable. This motivates a natural question: can tactile signals be inferred purely from vision?
To address this, we introduce \emph{EgoTac}, a generalizable model that predicts rich tactile information directly from egocentric human videos. EgoTac is trained on a unified corpus of over 5.7M image--tactile pairs, covering both continuous force measurements and binary contacts. By learning from this diverse dataset, EgoTac captures nuanced touch dynamics across varied interactions.
Experiments demonstrate strong performance: in-domain prediction achieves an average force error below 0.06N. On out-of-domain contact prediction benchmarks, EgoTac consistently outperforms the state-of-the-art contact estimator. It also captures the rise and fall patterns of real tactile data and enables zero-shot predictions on unconstrained real-world videos. Scaling analyses further reveal that both data diversity and volume improve performance steadily.
Overall, EgoTac provides a scalable pathway to extract tactile priors from egocentric human videos, enabling broadly applicable tactile-aware robot learning. Related resources can be found at https://egotac.github.io/.
\end{abstract}
    
\section{Introduction}
\label{sec:intro}

Dexterity is central to physical intelligence and robot learning~\cite{intelligence2025pi_, bjorck2025gr00t}, while touch is essential to dexterity. 
Everyday skills such as wiping a vase, opening a door, or inserting a socket require robots not only to perceive the scene, but also to continuously regulate physical contact: where to touch, how much pressure to apply, and when to release as the interaction evolves. However, learning such tactile-aware behavior requires large amounts of interaction data that capture rich manipulation patterns. 
Since collecting this data directly on robots remains difficult and costly, researchers increasingly turn to human data as a scalable source of manipulation experience~\cite{zheng2026egoscale,luo2025being,kareer2025emergence}. 

Large-scale egocentric human datasets~\cite{hoque2025egodex,zhan2024oakink2,liu2024taco,fan2023arctic} capture diverse  behaviors and offer a promising substrate for learning physical skills, but they typically stop at pixels, hand poses, and object trajectories. 
The tactile variables are absent largely due to the difficulty of capturing tactile signals at scale. Unlike vision, which can be recorded passively and remotely, touch must be measured at the physical contact interface, making data collection sensitive to sensor placement, calibration, spatial coverage, durability, and wearability. 

This difficulty motivates a complementary approach: instead of collecting touch everywhere, can we recover tactile information from the human videos that already exist? 
If dense hand tactile states can be inferred from in-the-wild egocentric videos, existing visual datasets of human manipulation can be augmented with contact and force-like physical supervision without requiring tactile sensors at collection time. 
In this paper, we study this novel \textbf{vision-to-tactile prediction problem}, aiming to turn large-scale human video corpora into a scalable source of tactile supervision.

\begin{figure*}[t]
\begin{center}
\includegraphics[width=1.0\linewidth]{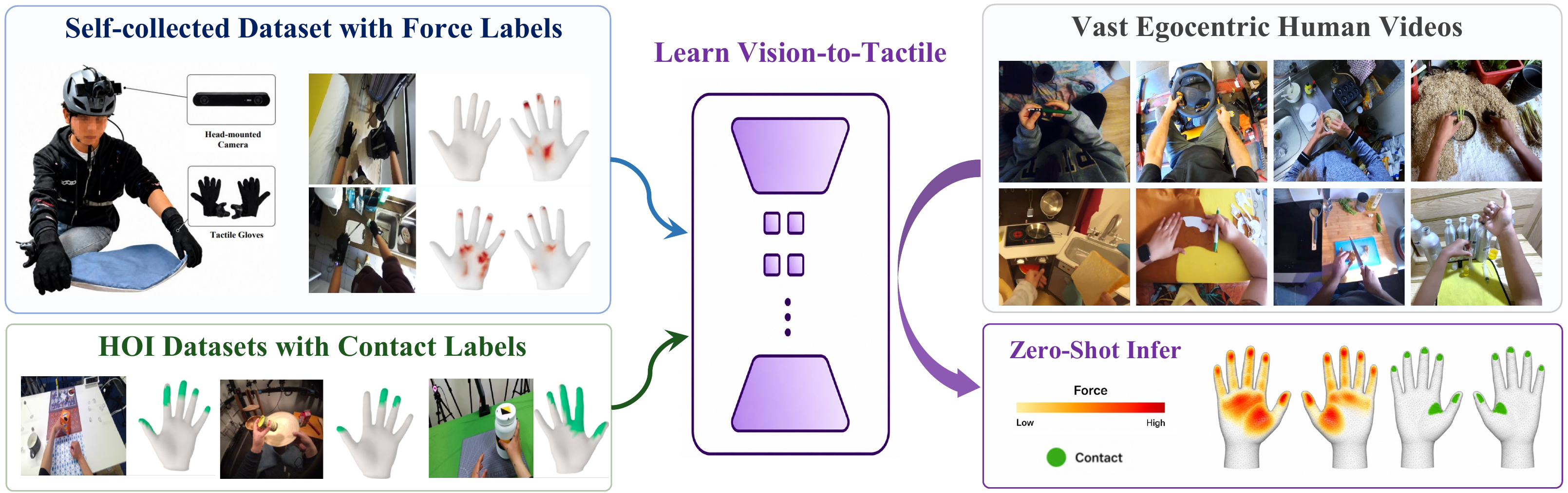}
\end{center}
\vspace{-2.5mm}
\caption{
\textbf{Overview of EgoTac.} EgoTac unifies self-collected tactile data with contact-labeled egocentric hand-object interaction datasets into a shared MANO-aligned representation, and learns a vision-to-tactile model that predicts dense force and contact from egocentric RGB videos. The learned mapping enables zero-shot tactile labeling of vast in-the-wild human video corpora, providing scalable physical supervision for robot learning and beyond.}
\label{fig:teaser}
\vspace{-0.4cm}
\end{figure*}


To address this, we introduce \textbf{\emph{EgoTac}, the first generalizable model for vision-to-dense tactile prediction from in-the-wild egocentric human videos.} 
To support unified training and evaluation, we first construct \emph{EgoTac-Dataset}, a large-scale dataset that contains over \textbf{5.7 million} image-tactile pairs. It combines our self-collected egocentric dataset \emph{EgoTac-SC} and eight prior hand-object interaction datasets, covering both single-hand and bimanual interactions across a wide range of real-world tasks and scenes. To fully use these heterogeneous sources with differing spatial coverage, we map all annotations into a shared MANO~\cite{romero2017embodied} hand representation~\cite{zhang2025unitachand}.


Building on this corpus, we train \emph{EgoTac} as a unified and generalizable vision-to-touch model. 
\emph{EgoTac} follows a concise encoder--decoder architecture and jointly predicts dense continuous tactile values and contact classification labels from temporal vision input. 
With enough data scale for training, \emph{EgoTac} learns transferable vision-to-touch knowledge across diverse tasks, scenes, and annotation types. 
An overview of the framework is shown in Figure~\ref{fig:teaser}.

Experiments demonstrate EgoTac's effectiveness along three complementary axes. First, in-domain evaluation shows accurate force prediction ($\text{MAE}_\text{act}$ < 0.06\,N) and strong contact estimation (F1 > 0.70). Second, \textbf{out-of-domain and zero-shot transfer} experiments highlight its robust generalization. EgoTac consistently outperforms state-of-the-art methods on unseen contact benchmarks (OAKINK2~\cite{zhan2024oakink2}, FPHA~\cite{garcia2018first}) and preserves temporal force dynamics on OpenTouch~\cite{song2025opentouch}. Furthermore, it enables \textbf{plausible zero-shot tactile predictions on unconstrained datasets}, including EgoDex~\cite{hoque2025egodex}, EPIC-KITCHENS~\cite{damen2018scaling}, and Ego4D~\cite{grauman2022ego4d}. Third, data scaling analysis confirms consistent performance improvements as the training corpus grows. Collectively, these findings show that EgoTac delivers accurate in-domain tactile predictions and transferable contact and tactile-dynamics estimates across diverse out-of-domain settings.

Our contributions can be summarized as follows:
\begin{itemize}
\item \textbf{A unified visual-tactile dataset for egocentric hand-object interaction.} 
We introduce a large-scale MANO-aligned dataset that merges our newly collected EgoTac-SC dataset with eight prior hand-object interaction datasets, yielding over 5.7~M samples with real force or dense contact annotations.

\item \textbf{The first in-the-wild tactile prediction model from egocentric vision.}
We propose EgoTac, the first model to jointly predict dense tactile force and contact over both hands from egocentric RGB videos. EgoTac enables effective training from heterogeneous data sources and exhibits clear data-scaling benefits across diverse real-world scenarios.

\item \textbf{Comprehensive empirical evaluation of tactile prediction and generalization.} 
We evaluate EgoTac on in-domain prediction, out-of-domain benchmarks, and zero-shot transfer to in-the-wild egocentric scenes. Results demonstrate consistent improvements over prior work and robust tactile prediction under diverse scenes.

\end{itemize}

\section{Related Work}
\label{sec:related_works}

\begin{table*}[t]
\centering
\small
\caption{\textbf{Comparison of egocentric hand-object interaction datasets.} We summarize the key properties of existing representative egocentric datasets and our newly collected \emph{EgoTac-SC} dataset.}
\scalebox{0.85}{\begin{tabular}{l|cc|cccc|cc} \toprule
Dataset & \# Frames & \# Clips & Bimanual & Hand Pose & Contact & Real Force & \# Envs & In-the-wild \\
\midrule
Ego4D(HOI)~\cite{grauman2022ego4d} & 21M & 89K & \(\checkmark\) & \(\times\) & \(\times\) & \(\times\) & 900 & \(\checkmark\) \\
EPIC-KITCHENS~\cite{damen2018scaling} & 11.5M & 432 & \(\checkmark\) & \(\times\) & \(\times\) & \(\times\) & 32 & \(\checkmark\) \\
EgoPAT3D~\cite{li2022egocentric} & 1.1M & 150 & \(\times\) & Estimation & \(\times\) & \(\times\) & 15 & \(\checkmark\) \\
HOI4D~\cite{liu2022hoi4d} & 2.4M & 4K & \(\times\) & Estimation & \(\times\) & \(\times\) & 610 & \(\checkmark\) \\
FPHA~\cite{garcia2018first} & 105K & 1.2K & \(\checkmark\) & MoCap & \(\times\) & \(\times\) & 3 & \(\times\) \\
TACO~\cite{liu2024taco} & 5.2M & 2.5K & \(\checkmark\) & MoCap & \(\times\) & \(\times\) & 1 & \(\times\) \\
HOT3D~\cite{banerjee2025hot3d} & 1.5M & 425 & \(\checkmark\) & MoCap & \(\times\) & \(\times\) & 4 & \(\times\) \\
HO-Cap~\cite{wang2024ho} & 656K & 64 & \(\checkmark\) & Estimation & \(\times\) & \(\times\) & 1 & \(\times\) \\
OakInk2~\cite{zhan2024oakink2} & 4M & 627 & \(\checkmark\) & MoCap & \(\times\) & \(\times\) & 1 & \(\times\) \\
EgoDex~\cite{hoque2025egodex} & 90M & 338K & \(\checkmark\) & Estimation & \(\times\) & \(\times\) & 1 & \(\times\) \\
H2O~\cite{kwon2021h2o} & 571K & 164 & \(\checkmark\) & Estimation & Analytical & \(\times\) & 3 & \(\times\) \\
ARCTIC~\cite{fan2023arctic} & 2.1M & 339 & \(\checkmark\) & MoCap & Analytical & \(\times\) & 1 & \(\times\) \\
ActionSense~\cite{delpreto2022actionsense} & 1.4M & 574 & \(\checkmark\) & Glove & Pressure & \(\checkmark\) & 1 & \(\times\) \\
PressureVisionDB~\cite{grady2022pressurevision} & 3.5M & 12K & \(\times\) & \(\times\) & Pressure & \(\checkmark\) & 1 & \(\times\) \\
EgoPressure~\cite{zhao2025egopressure} & 4.3M & 1.3K & \(\times\) & Estimation & Pressure & \(\checkmark\) & 1 & \(\times\) \\
OpenTouch~\cite{song2025opentouch} & 550K & 2.9K & \(\times\) & Glove & Pressure & \(\checkmark\) & 14 & \(\checkmark\) \\
FEEL~\cite{dessalene2026feel} & 3M & - & \(\checkmark\) & Estimation & Pressure & \(\checkmark\) & 1 & \(\checkmark\) \\
EgoTac-SC(Ours) & 3.9M & 857 & \(\checkmark\) & Glove & Pressure & \(\checkmark\) & 24 & \(\checkmark\) \\
\bottomrule
\end{tabular}}
\vspace{-0.3cm}
\label{tab:dataset_comparison}
\end{table*}

\subsection{Egocentric Datasets for Hand-Object Interaction}
Egocentric datasets are essential for computer vision~\cite{liu2024easyhoi, li2026egocentric} and robot learning~\cite{kareer2025egomimic, yuan2025motiontrans, kareer2025emergence, zheng2026egoscale}. Early foundational collections primarily focused on visual observations~\cite{grauman2022ego4d, damen2018scaling}, enabling large-scale understanding of daily activities but lacking fine-grained physical interaction annotations required for dexterous manipulation. To address this, subsequent multimodal datasets~\cite{li2022egocentric,liu2022hoi4d, liu2024taco, garcia2018first, banerjee2025hot3d, zhan2024oakink2, wang2024ho, hoque2025egodex, fan2023arctic, kwon2021h2o} introduced 3D hand and object poses. However, these datasets typically rely on analytical contact labels inferred computationally from mesh proximity rather than genuine physical contact. 
More recently, efforts have shifted towards incorporating direct tactile signals~\cite{zhao2025egopressure, song2025opentouch, dessalene2026feel}. Yet, these resources often face significant limitations: 
they are typically constrained to specific interaction surfaces~\cite{grady2022pressurevision, zhao2025egopressure}, limited in scale~\cite{song2025opentouch}, or lacking real-world scene diversity~\cite{delpreto2022actionsense, dessalene2026feel}.
To overcome these critical bottlenecks, we introduce a large-scale, unified visual-tactile dataset yielding over 5.7~M samples. By aggregating 8 well-established egocentric datasets featuring analytical mesh-based contacts with our newly collected RGB-force dataset \emph{EgoTac-SC}, we provide a standardized MANO-aligned format that bridges the gap between simulated contacts and real-world high-resolution force signals. We summarize and compare these representative egocentric datasets in Table~\ref{tab:dataset_comparison}.

\subsection{Tactile Prediction from Vision}
Vision-based tactile prediction has steadily progressed from localized texture recognition to the inference of full-hand force distributions. Early investigations largely relied on optical-tactile sensors like GelSight~\cite{yuan2017gelsight} to build cross-modal mappings from RGB images~\cite{li2019connecting, gao2023objectfolder}. However, these approaches primarily captured geometric and textural cues, neglecting the complex force dynamics inherent in dexterous manipulation. On the other hand, contact-centric methods~\cite{brahmbhatt2019contactdb, grady2021contactopt, jung2025learning} successfully employ thermal imaging or mesh-distance heuristics to localize interaction areas, but they fail to output the continuous force values critical for precision tasks.
To bypass the severe scarcity of dense ground-truth tactile labels, simulation-based methods~\cite{ehsani2020use, hanai2023force} often supervise tactile inference with physics engines, but they suffer a persistent sim-to-real gap. While recent efforts have attempted real-world force prediction, they are predominantly confined to controlled laboratory setups~\cite{zhao2025egopressure, li2023learning, grady2022pressurevision, grady2024pressurevision++} or specifically tailored to localized regions such as fingertips~\cite{pham2015towards, fallahinia2022real, chen2020estimating, pham2017hand}. By leveraging our proposed unified dataset, this work takes a substantial step towards achieving robust, full-hand tactile prediction from egocentric vision across unconstrained daily environments.

\section{Unified Dataset for Visual-Tactile Learning}
\label{sec:dataset}

\begin{figure*}[t]
\begin{center}
\includegraphics[width=1.0\linewidth]{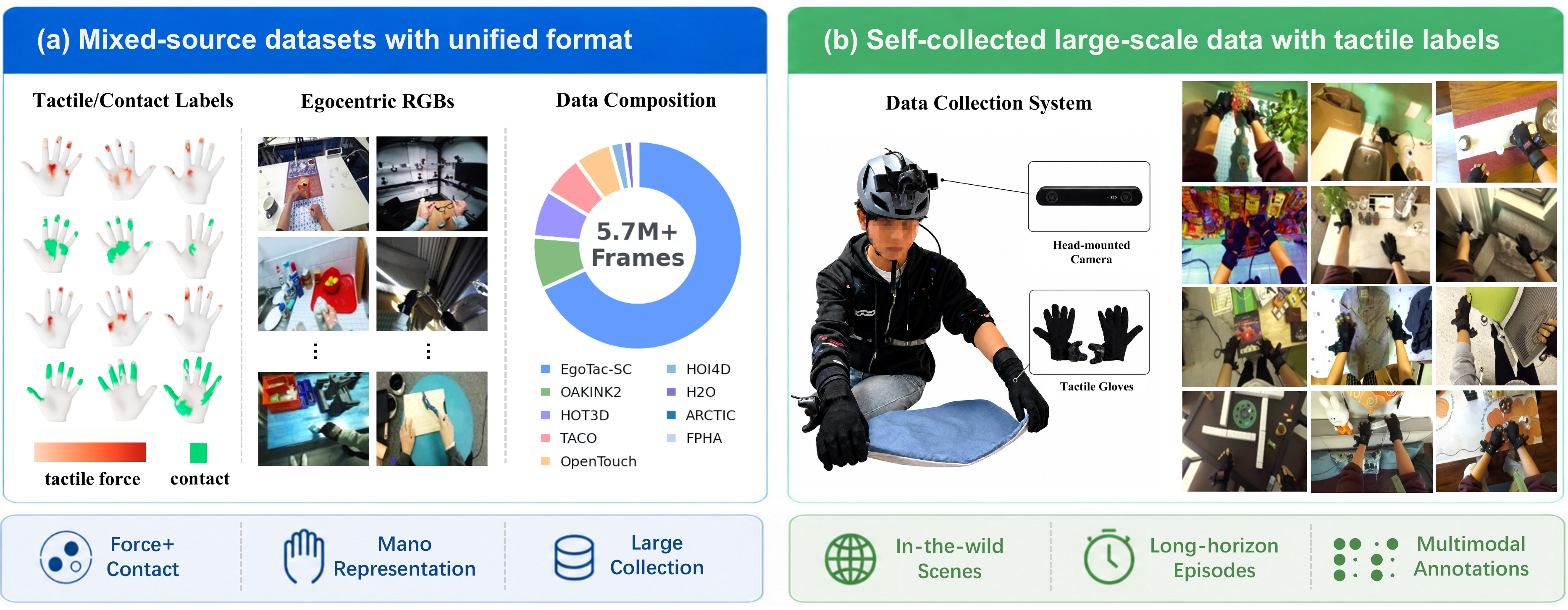}
\end{center}
\vspace{-2.5mm}
\caption{
\textbf{Dataset overview.} (a) We aggregate mixed-source hand-object interaction datasets containing real tactile measurements or mesh-based contact labels, all unified under a shared MANO topology. (b) The largest portion of our corpus is the self-collected \emph{EgoTac-SC} dataset, which utilizes a custom tactile sensing hardware setup to capture diverse long-horizon manipulation data.}
\label{fig:dataset_overview}
\vspace{-0.4cm}
\end{figure*}

\subsection{EgoTac Dataset Overview}
We aggregate 9 datasets into a unified visual-tactile dataset with more than 5.7M samples: our self-collected EgoTac-SC dataset and 8 established hand-object interaction datasets. Figure~\ref{fig:dataset_overview}(a) summarizes the dataset composition and annotation types. We next detail the unified format, custom collection process, and dataset integration.

\subsection{Unified Data Format}
\label{sec:dataset_format}
We use a unified annotation format for all datasets. Let $t$ denote the number of time steps and let $V=778$ be the number of vertices of one MANO hand~\cite{romero2017embodied}. For two-hand interactions, the total number of vertices is $2V$. For each sequence, we represent annotations as
\begin{equation}
\mathbf{F} \in \mathbb{R}^{t \times (2V) \times 1}, \qquad
\mathbf{C} \in \{0,1\}^{t \times (2V) \times 1}.
\end{equation}
Here, $\mathbf{F}$ stores continuous tactile intensity values, and $\mathbf{C}$ stores binary contact labels (0 for non-contact, 1 for contact). This representation enables seamless joint training with tactile-rich and contact-only data sources.

\subsection{Custom Tactile Data Collection}
As a core component of the unified dataset, we collect a new tactile dataset, \emph{EgoTac-SC}, that captures high-resolution force signals during natural daily interactions. Participants wear custom flexible tactile gloves with 264 force sensors together with a commodity hand-tracking glove for 3D hand pose capture. Visual context is recorded by a head-mounted egocentric camera at 30 FPS (hardware setup shown in Figure~\ref{fig:dataset_overview}(b)). Participants perform more than 70 contact-rich manipulation tasks (e.g., folding clothes, wiping tables, and opening doors) in 24 diverse environments, including kitchens, offices and more. During post-processing, raw tactile readings are filtered and mapped to MANO vertices to match the unified format in Sec.~\ref{sec:dataset_format}. Additional details are shown in Appendix~\ref{sec:supp_egotac_sc}.

\subsection{Integration of Existing HOI Datasets}
To further improve scale and coverage, we process and integrate existing egocentric HOI datasets. These include OpenTouch~\cite{song2025opentouch}, which provides real tactile measurements but mainly single-hand interactions, and contact-only datasets including HOT3D~\cite{banerjee2025hot3d}, TACO~\cite{liu2024taco}, HOI4D~\cite{liu2022hoi4d}, H2O~\cite{kwon2021h2o}, ARCTIC~\cite{fan2023arctic}, OAKINK2~\cite{zhan2024oakink2}, and FPHA~\cite{garcia2018first}. For these contact-only datasets with hand and object meshes, we derive vertex-level contact labels following~\cite{jung2025learning}: a MANO vertex is marked as contact when its minimum distance to the object mesh is below a threshold. More processing details are provided in Appendix~\ref{sec:supp_other_datasets}.

\section{Vision-to-Tactile Prediction Framework}
\label{sec:method}

\begin{figure*}[t]
\begin{center}
\includegraphics[width=1.0\linewidth]{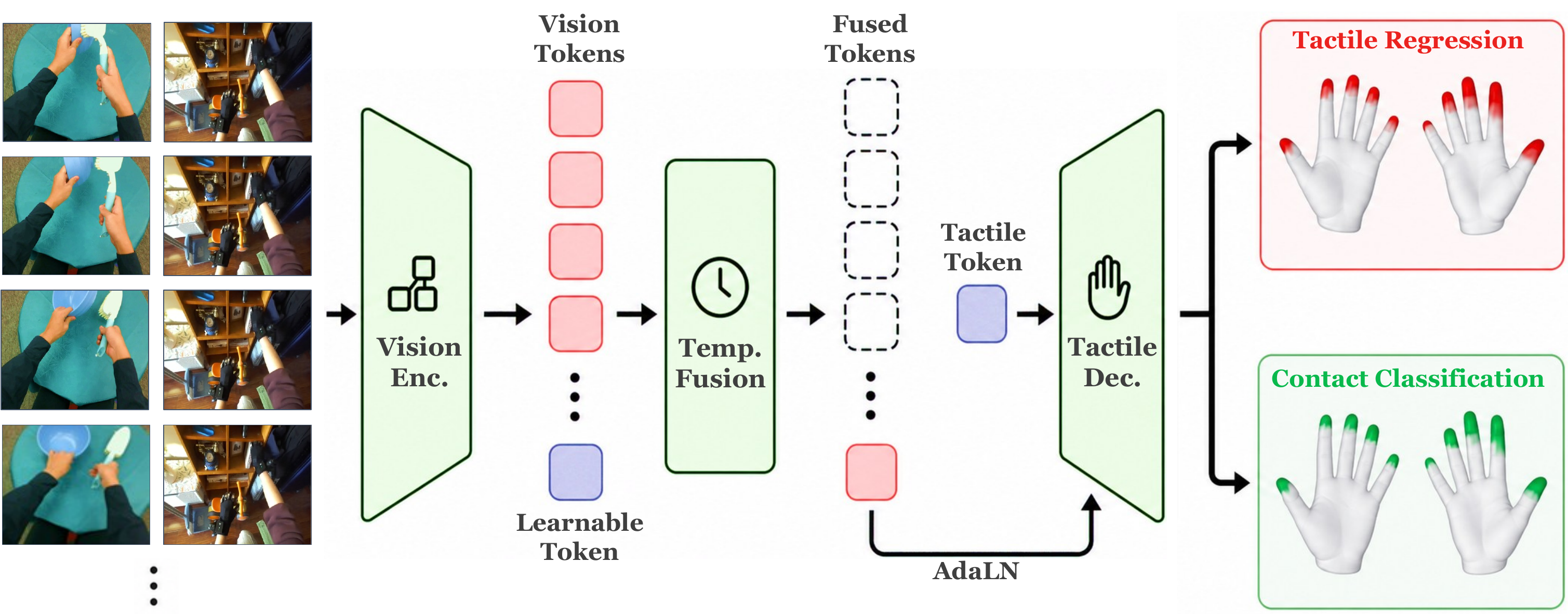}
\end{center}
\vspace{-2.5mm}
\caption{
\textbf{Model architecture of EgoTac.} Given a short history of egocentric RGB frames, a pretrained vision encoder computes spatial-semantic tokens, and a temporal fusion module aggregates global context over time. An AdaLN-based tactile decoder then jointly conducts dense force regression and contact classification on the shared MANO hand topology.}
\label{fig:pipeline}
\vspace{-0.4cm}
\end{figure*}

Predicting dense tactile states from unconstrained egocentric vision is fundamentally challenging. Prior vision-based methods were predominantly confined to controlled laboratory setups~\cite{zhao2025egopressure}, focused only on localized regions (e.g., fingertips)~\cite{pham2015towards}, or relied purely on analytical mesh-distance heuristics~\cite{jung2025learning}. To address these bottlenecks, we introduce a unified prediction framework that leverages the large-scale heterogeneous dataset described in Sec.~\ref{sec:dataset}. Our key design is a joint force-contact training scheme that cohesively aggregates continuous force measurements with discrete contact annotations into a shared MANO representation. We detail our problem formulation in Sec.~\ref{sec:problem}, model architecture in Sec.~\ref{sec:architecture}, and training strategy in Sec.~\ref{sec:training}.

\subsection{Problem Formulation}
\label{sec:problem}
We aim to infer dense tactile states of both hands from egocentric RGB observations. Given an egocentric RGB clip $\mathbf{I}_{t-T+1:t}\in\mathbb{R}^{T\times 3\times H\times W}$ with an observation horizon $T$, our model concurrently predicts continuous tactile force fields $\hat{\mathbf{F}}$ and discrete contact states $\hat{\mathbf{C}}$ over a prediction horizon $K$, all anchored on the MANO hand mesh~\cite{romero2017embodied}:
\begin{equation}
    (\hat{\mathbf{F}}_{t:t+K-1}, \hat{\mathbf{C}}_{t:t+K-1})=f_\theta(\mathbf{I}_{t-T+1:t}), \qquad
    \hat{\mathbf{F}}, \hat{\mathbf{C}}\in\mathbb{R}^{K\times 2V},
\end{equation}
where $V=778$ designates the number of vertices on the MANO mesh, and the factor of two accounts for both the left and right hands. Specifically, $\hat{\mathbf{F}}$ denotes the scalar normal force at each vertex, while $\hat{\mathbf{C}}$ represents the corresponding per-vertex contact probability logits.

\subsection{EgoTac Model Architecture}
\label{sec:architecture}
The architecture of EgoTac is illustrated in Figure~\ref{fig:pipeline}. Our model consists of a vision encoder, a temporal fusion module, and an AdaLN-based tactile decoder. Each egocentric RGB frame within the observation horizon is encoded by a pretrained DINOv2 ViT~\cite{oquab2023dinov2} into spatial-semantic vision tokens. To aggregate information over time, we use a learnable temporal token that attends to the visual sequence via cross-attention, producing a global context vector $z \in \mathbb{R}^{d}$.

Conditioned on $z$, the tactile decoder performs direct prediction in a single forward pass. It maintains a set of learnable tactile query tokens
$q \in \mathbb{R}^{K \times d}$, where each token corresponds to tactile estimations at one prediction step. These tokens are iteratively updated through Transformer layers modulated by Adaptive Layer Normalization (AdaLN):
\begin{equation}
\mathrm{AdaLN}_{\ell}(h; z) = (1 + \gamma_{\ell}(z)) \odot h + \beta_{\ell}(z),
\end{equation}
where $h \in \mathbb{R}^{K \times d}$ denotes the token features at layer $\ell$, and the scale $\gamma_{\ell}(z)$ and shift $\beta_{\ell}(z)$ are dynamically generated from the context vector $z$ and broadcast over the token dimension. This conditioning mechanism preserves computational efficiency while enabling the temporal visual context to guide predictions over the entire MANO topology. 

\subsection{Force-Contact Joint Training}
\label{sec:training}
Dense tactile inference requires predicting both where interaction occurs and how strong it is. We therefore attach two task-specific heads to the shared decoder: (1) a tactile regression head for continuous force values, and (2) a contact classification head for per-vertex contact logits. This dual formulation enables mixed-dataset learning: force-supervised samples provide both targets, while contact-only samples provide contact targets without force magnitudes.

Specifically, our training set mixes the force-supervised egocentric tactile source with geometry-derived contact-only datasets. For force-supervised samples, contact labels can be derived by thresholding the continuous force; for contact-only samples, they come from mesh-based collision proximity. For each sample, the dataloader provides a force-valid mask $\mathbf{M}^{F}$ and a contact-valid mask $\mathbf{M}^{C}$ for coherent supervision even with missing annotations. The total objective is defined as
\begin{equation}
    \mathcal{L}
    = \lambda_F\mathcal{L}_{F}
    + \lambda_C\mathcal{L}_{C}
    + \lambda_A\mathcal{L}_{A}
    + \lambda_{\mathrm{cons}}\mathcal{L}_{\mathrm{cons}} .
\end{equation}
Here, $\mathcal{L}_F$ is an MSE loss on normalized tactile values, applied only where continuous force supervision is available. The contact loss $\mathcal{L}_C$ is a BCE loss over contact-valid vertices with positive class reweighting. To avoid trivial all-zero predictions, we apply an active-region loss $\mathcal{L}_A$ as a Smooth-$L_1$ penalty on vertices with active ground-truth force. Finally, the consistency loss $\mathcal{L}_{\text{cons}}$ aligns contact probabilities derived from the force branch with detached contact-head predictions via BCE. Formal definitions are provided in Appendix~\ref{sec:supp_impl_loss}.

For contact-only samples, $\mathbf{M}^{F}=0$, so neither $\mathcal{L}_F$
nor $\mathcal{L}_A$ contributes. Nevertheless, $\mathcal{L}_C$ updates the
shared encoder and decoder, while $\mathcal{L}_{\mathrm{cons}}$ constrains
where and when the force branch is active. Contact thus supplies a structural
constraint, while EgoTac-SC provides force magnitudes. We examine these
complementary roles in Section~\ref{sec:heterogeneous_supervision}.

\subsection{Implementation Details}
\label{sec:implementation}
We optimize the full model end-to-end with AdamW~\cite{loshchilov2018decoupled} on 4 A10 GPUs, using a total batch size of 128, a cosine-decayed learning rate schedule from 1e-4 to 1e-5 and 5\% warm-up of the total training steps. Unless otherwise noted, the vision encoder is fine-tuned jointly with the decoder using a reduced learning-rate multiplier of 0.1, and we maintain an EMA model for evaluation. Additional details, including decoder depth and width, dataset sampling, loss weights, and augmentation settings, are provided in the Appendix~\ref{sec:supp_impl_hparams}.

\section{Experiments}
\label{sec:experiments}

\begin{figure*}[t]
\begin{center}
\includegraphics[width=1\linewidth]{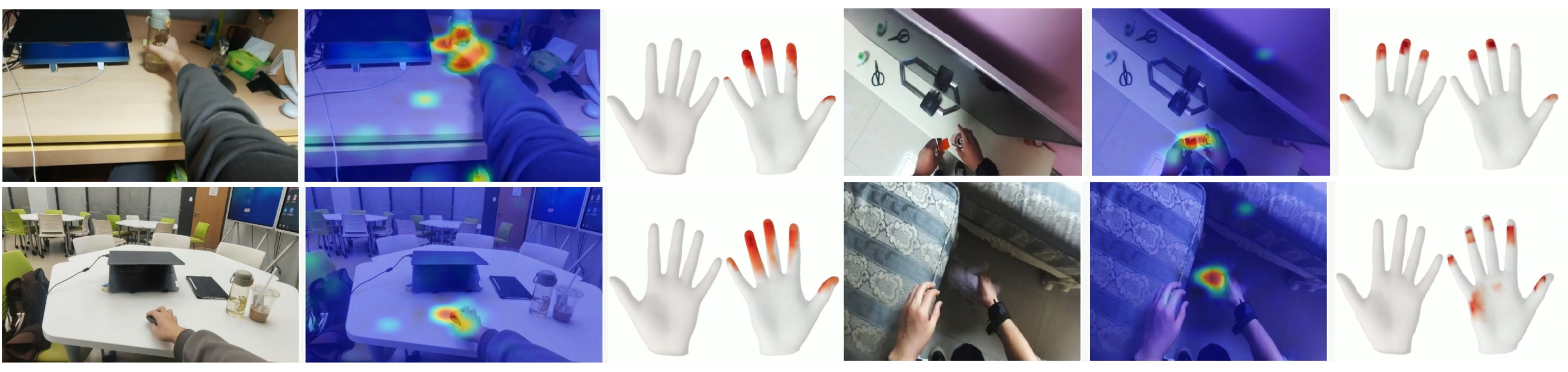}
\end{center}
\vspace{-2.5mm}
\caption{
\textbf{Attention maps from EgoTac.} Across casual in-the-wild videos, EgoTac emphasizes hand-object interaction regions that are informative for tactile prediction.}
\label{fig:attention}
\vspace{-0.4cm}
\end{figure*}

\subsection{Experimental Overview}

We evaluate EgoTac through four research questions:
\begin{itemize}
	\item \textbf{In-domain performance and ablations:} On seen domains, how accurately does EgoTac predict continuous force and binary contact, and how much do key design choices contribute? (Section~\ref{sec:in_domain})
	\item \textbf{Out-of-domain generalization:} Under domain shift, how does EgoTac compare with prior SOTA contact estimators, and does it preserve force dynamics on real tactile data? Further, is it possible to zero-shot infer tactile labels on diverse HOI scenes? (Section~\ref{sec:ood_eval})
    \item \textbf{Data scaling effects:} How do training data scale and diversity affect EgoTac’s prediction performance? (Section~\ref{sec:datascaling})
    \item \textbf{Supervision sources:} How do measured-force and contact-only data contribute to contact and force prediction? (Section~\ref{sec:heterogeneous_supervision})
\end{itemize}

\paragraph{Datasets.}
Training uses EgoTac-SC together with HOI4D~\cite{liu2022hoi4d}, H2O~\cite{kwon2021h2o}, HOT3D~\cite{banerjee2025hot3d}, ARCTIC~\cite{fan2023arctic} and TACO~\cite{liu2024taco}. For in-domain evaluation, each dataset is split 9:1, and we report results on the held-out 10\%. For out-of-domain (OOD) evaluation, we use OpenTouch~\cite{song2025opentouch}, OAKINK2~\cite{zhan2024oakink2}, and FPHA~\cite{garcia2018first} without further splitting.

\paragraph{Baselines.}
For in-domain evaluation, we mainly compare EgoTac with EgoTac-i, an image-based variant that removes temporal context.
For out-of-domain contact benchmarks, we compare against three representative RGB-based 3D contact estimators: BSTRO~\cite{huang2022capturing}, DECO~\cite{tripathi2023deco}, and HACO~\cite{jung2025learning}.
BSTRO and DECO predict vertex-level 3D contact from RGB observations and are therefore directly comparable to our MANO-aligned contact prediction setting. HACO is the state-of-the-art dense contact estimator, which first uses a hand detector~\cite{potamias2025wilor} to crop the right hand and then performs contact inference only on the right hand.

\paragraph{Metrics.}
For in-domain continuous tactile prediction, we report global MAE, active-region MAE ($\text{MAE}_{\textbf{act}}$) and temporal Pearson correlation between predicted and ground-truth force. Unless otherwise stated, active vertices are defined by ground-truth force $>0.05\,\mathrm{N}$. For contact benchmarks, we report Precision, Recall, F1, IoU, and AUROC from contact logits. For the OOD force-sensitive benchmark, we report relative force dynamics metrics including cosine similarity, rise F1 and fall F1. Additionally, we provide metric details in Appendix~\ref{sec:supp_metrics} and extensive frequency-domain diagnostics in Appendix~\ref{sec:supp_more_results}.

\begin{table}[t]
\centering
\small
\setlength{\tabcolsep}{9pt}
\caption{\textbf{In-domain tactile evaluation and ablation results.} Averaged metrics on the EgoTac-SC test set are reported. Ablations validate effects of  the temporal visual context, the active-region loss, and explicit force supervision.}
\label{tab:in_domain_tactile}
\begin{tabular}{lccc}
\toprule
Method & MAE $\downarrow$ & $\text{MAE}_{\textbf{act}}$ $\downarrow$ & Temporal Pearson $\uparrow$  \\
\midrule
\textbf{EgoTac} & \textbf{0.004} & \textbf{0.056} & \textbf{0.506} \\
w/o temporal context & 0.005 & \textbf{0.056} & 0.463 \\
w/o active-region loss & \textbf{0.004} & 0.097 & 0.482 \\
w/o force supervision & 0.322  & 0.108  & 0.240 \\
\bottomrule
\end{tabular}
\vspace{-0.2cm}
\end{table}


\begin{table}[t]
\centering
\small
\setlength{\tabcolsep}{4pt}
\caption{\textbf{In-domain contact evaluation.} Results are averaged across timestamps on all test splits of each component dataset. EgoTac maintains strong contact localization accuracy despite the additional force regression objectives.}
\label{tab:id_contact}
\begin{tabular}{llccccc}
\toprule
Benchmark & Method & AUROC $\uparrow$ & IoU $\uparrow$ & Precision $\uparrow$ & Recall $\uparrow$ & F1 $\uparrow$ \\
\midrule
\multirow{2}{*}{HOT3D~\cite{banerjee2025hot3d}} 
& EgoTac-i & \textbf{0.976} & 0.587 & 0.656 & \textbf{0.771} & 0.702 \\
& \textbf{EgoTac} & \textbf{0.976} & \textbf{0.593} & \textbf{0.667} & 0.767 & \textbf{0.706} \\
\midrule
\multirow{2}{*}{H2O~\cite{kwon2021h2o}} 
& EgoTac-i & \textbf{0.989} & 0.647 & 0.725 & 0.846 & 0.777 \\
& \textbf{EgoTac} & \textbf{0.989} & \textbf{0.743} & \textbf{0.804} & \textbf{0.904} & \textbf{0.850} \\
\midrule
\multirow{2}{*}{ARCTIC~\cite{fan2023arctic}} 
& EgoTac-i & 0.972 & \textbf{0.676} & 0.716 & \textbf{0.905} & 0.790 \\
& \textbf{EgoTac} & \textbf{0.977} & 0.669 & \textbf{0.732} & 0.897 & \textbf{0.795} \\
\bottomrule
\end{tabular}
\vspace{-0.2cm}
\end{table}

\subsection{In-Domain Performance and Ablations}
\label{sec:in_domain}
\paragraph{In-domain tactile-based evaluation.}
We first evaluate calibrated force prediction on held-out EgoTac-SC episodes. Because non-contact vertices dominate the MANO surface, global averages can underestimate errors on physically meaningful regions; therefore, we prioritize $\text{MAE}_{\textbf{act}}$ and use global MAE and temporal correlation as complementary signals.
Table~\ref{tab:in_domain_tactile} shows three main trends. First, incorporating temporal visual context improves temporal consistency.  
Second, adding the active-region loss enhances force fidelity in interacting areas.  
Third, relying on contact-only supervision is insufficient for realistic force estimation, confirming that contact logits do not directly encode the force scale.

\begin{table}[t]
\centering
\small
\setlength{\tabcolsep}{4pt}
\caption{\textbf{Out-of-domain contact evaluation.} Results are averaged across timestamps on OAKINK2~\cite{zhan2024oakink2} and FPHA~\cite{garcia2018first}. EgoTac yields superior performance compared with previous methods.}
\label{tab:ood_contact}
\begin{tabular}{llccccc}
\toprule
Benchmark & Method & AUROC $\uparrow$ & IoU $\uparrow$ & Precision $\uparrow$ & Recall $\uparrow$ & F1 $\uparrow$ \\
\midrule
\multirow{4}{*}{Oakink2~\cite{zhan2024oakink2}} 
& BSTRO~\cite{huang2022capturing} & 0.537 & 0.076 & 0.172 & 0.139 & 0.134 \\
& DECO~\cite{tripathi2023deco} & 0.655 & 0.144 & 0.291 & 0.329 & 0.243 \\
& HACO~\cite{jung2025learning} & 0.772 & 0.259 & 0.356 & \textbf{0.568} & 0.400 \\
& \textbf{EgoTac} & \textbf{0.844} & \textbf{0.285} & \textbf{0.498} & 0.441 & \textbf{0.435} \\
\midrule
\multirow{4}{*}{FPHA~\cite{garcia2018first}} 
& BSTRO~\cite{huang2022capturing} & 0.535 & 0.017 & 0.080 & 0.033 & 0.032 \\
& DECO~\cite{tripathi2023deco} & 0.635 & 0.041 & 0.161 & 0.092 & 0.075 \\
& HACO~\cite{jung2025learning} & 0.699 & 0.160 & 0.187 & \textbf{0.716} & 0.268 \\
& \textbf{EgoTac} & \textbf{0.758} & \textbf{0.178} & \textbf{0.304} & 0.401 & \textbf{0.290} \\
\bottomrule
\end{tabular}
\vspace{-0.2cm}
\end{table}

\begin{figure}[t]
\centering
\noindent
\begin{minipage}[t]{0.40\textwidth}
    \vspace{0pt} 
    \centering
    \scriptsize
    \includegraphics[width=\linewidth]{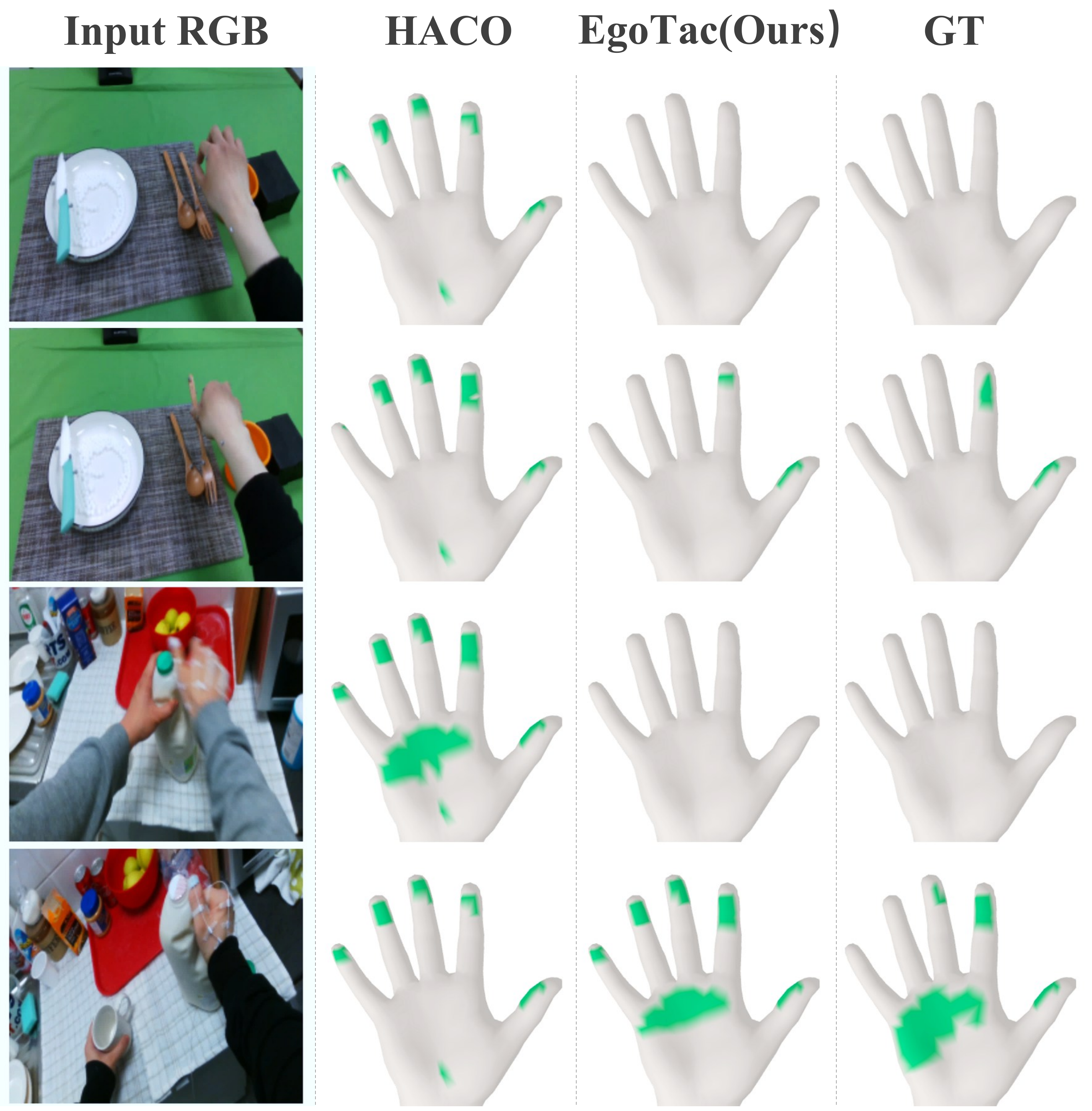}
    \vspace{-4mm}
    \captionof{figure}{\textbf{Qualitative comparison of OOD contact estimation (right-hand).} EgoTac produces cleaner contact status separation than HACO~\cite{jung2025learning} on OAKINK2~\cite{zhan2024oakink2} and FPHA~\cite{garcia2018first}.}
    \label{fig:ood_contact_vis}
\end{minipage}%
\hfill
\begin{minipage}[t]{0.535\textwidth}
    \vspace{0pt} 
    \centering
    \includegraphics[width=\linewidth]{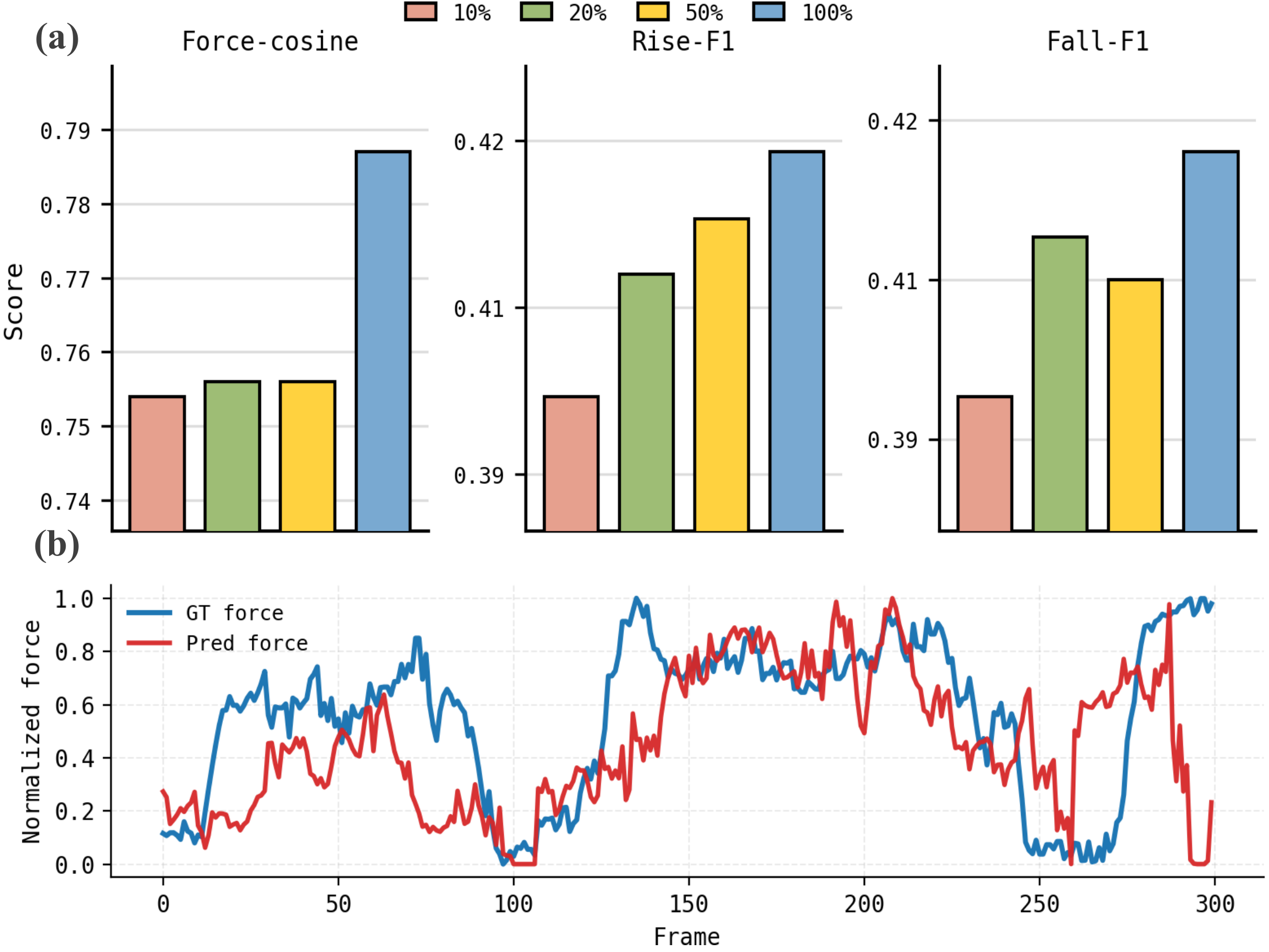}
      \vspace{-4mm}
      \caption{\textbf{Data scaling and OOD force prediction.} (a) Scaling training data volume improves force dynamics prediction on OpenTouch~\cite{song2025opentouch}. (b) Comparison of normalized total force over time, showing EgoTac effectively tracks ground-truth temporal trends.}
      \label{fig:ood_force_vis}
\end{minipage}
\vspace{-0.4cm}
\end{figure}

\paragraph{In-domain contact-based evaluation.}
Since EgoTac jointly predicts force and contact on a shared MANO topology, we also evaluate contact quality on in-domain HOI test splits. As shown in Table~\ref{tab:id_contact}, EgoTac maintains strong contact prediction performance (AUROC $>0.97$, F1 $>0.70$), indicating that improved force prediction is not achieved by sacrificing contact localization. Together, the force and contact evaluations provide a stricter in-domain validation of both physical intensity prediction and spatial interaction accuracy.

\subsection{Out-of-Domain Generalization}
\label{sec:ood_eval}
\paragraph{OOD contact-based benchmarks.}
Most public HOI datasets do not provide dense force supervision, so OOD evaluation is primarily contact-centric. 
We compare EgoTac with three representative RGB-based 3D contact
estimators, BSTRO~\cite{huang2022capturing}, DECO~\cite{tripathi2023deco}, and HACO~\cite{jung2025learning}, on OAKINK2~\cite{zhan2024oakink2} and FPHA~\cite{garcia2018first}.
As shown in Table~\ref{tab:ood_contact}, EgoTac achieves the best AUROC, IoU, Precision,
and F1 on both benchmarks.
HACO achieves higher Recall in some settings; however, this improvement largely stems from over-predicting contact regions, resulting in greater coverage at the cost of more false positives. Figure~\ref{fig:ood_contact_vis} qualitatively supports this trend: EgoTac yields cleaner separation between in-contact and out-of-contact states and produces more accurate contact maps, whereas HACO tends to predict contact even when no interaction occurs.

\paragraph{OOD force-sensitive benchmark on OpenTouch.}
To test OOD transfer beyond binary contact, we evaluate force dynamics on OpenTouch~\cite{song2025opentouch},  which uses a tactile sensing system different from EgoTac-SC. The public OpenTouch annotations do not provide the per-taxel physical calibration and effective sensing areas required for a reliable conversion to Newton level forces. Therefore, a direct absolute-force MAE in Newtons would require unsupported cross-sensor calibration assumptions.

Instead,  we normalize prediction and ground truth to $[0,1]$ using calibration-split statistics before evaluation.
This protocol measures whether EgoTac preserves relative tactile intensity and temporal structure under sensor and domain shift. 
In Figure~\ref{fig:ood_force_vis}(a), EgoTac reaches force-cosine similarity above 0.7 and rise/fall F1 above 0.4. Figure~\ref{fig:ood_force_vis}(b) further shows that predicted temporal force traces track the ground-truth trend. Overall, these results indicate that EgoTac can reliably predict continuous force dynamics even under out-of-domain conditions.

\begin{figure*}[t]
\begin{center}
\includegraphics[width=1.0\linewidth]{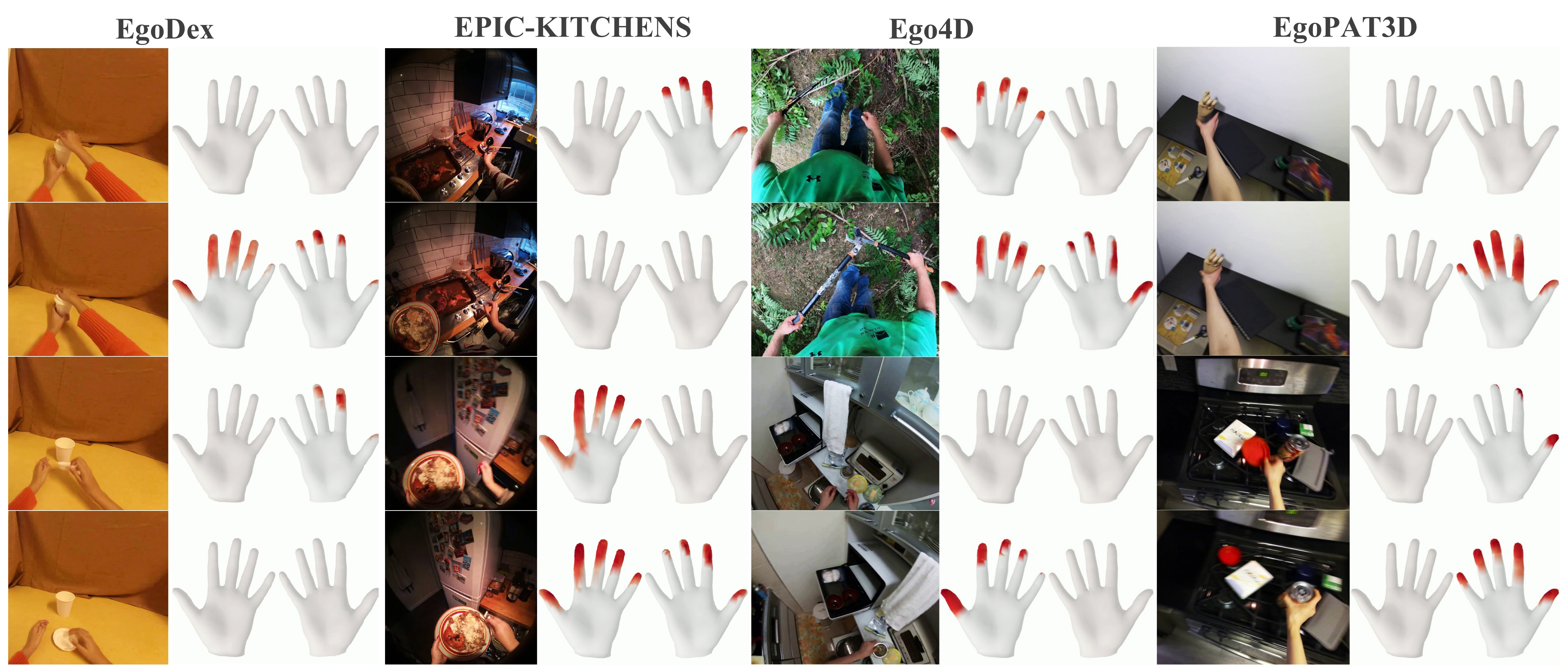}
\end{center}
\vspace{-0.1cm}
\caption{
\textbf{Qualitative results in diverse, in-the-wild scenes.} We visualize zero-shot tactile prediction results for both hands on clips from multiple egocentric datasets, including EgoDex~\cite{hoque2025egodex}, EPIC-KITCHENS~\cite{damen2018scaling}, Ego4D~\cite{grauman2022ego4d}, and EgoPAT3D~\cite{li2022egocentric}. EgoTac produces spatially coherent tactile patterns and smooth temporal transitions.}
\label{fig:vis_wild}
\vspace{-0.6cm}
\end{figure*}

\paragraph{Qualitative results on diverse HOI scenes.}
Figure~\ref{fig:vis_wild} shows qualitative predictions on clips from EgoDex~\cite{hoque2025egodex}, EPIC-KITCHENS~\cite{damen2018scaling}, Ego4D~\cite{grauman2022ego4d}, and EgoPAT3D~\cite{li2022egocentric}.
Across diverse real-world scenes, EgoTac produces spatially coherent tactile patterns and smooth temporal transitions.
In examples from EgoDex and EPIC-KITCHENS, EgoTac not only identifies which hand is actively engaged in manipulation, but also captures fine-grained contact and force changes over long-horizon episodes.
In EgoPAT3D examples, EgoTac further adapts its predicted tactile patterns according to the objects being interacted with, indicating sensitivity to object-specific geometry. Additionally, we use EgoTac to infer tactile from casual recorded videos, demonstrating the model's performance on custom data. Figure~\ref{fig:attention} presents attention maps in these self-recorded scenes, showing that EgoTac consistently focuses on hand-object interaction regions that determine tactile patterns.

\begin{figure}[t]
\centering
\noindent
\begin{minipage}[t]{0.52\textwidth}
    \vspace{10pt}
    \centering
    \small
    \setlength{\tabcolsep}{6pt}
    \captionof{table}{\textbf{Data diversity scaling on OOD contact estimation.} We increase the number of data sources used for training and report contact metrics on the OAKINK2~\cite{zhan2024oakink2} benchmark. Consistent improvements are shown with increasing data diversity.}
    \label{tab:ablation_datause}

    \begin{tabular}{lccc}
    \toprule
    Data Use & AUROC $\uparrow$ & IoU $\uparrow$ & F1 $\uparrow$ \\
    \midrule
    1 dataset & 0.603 & 0.065 & 0.120 \\
    4 datasets & 0.832 & 0.263 & 0.407 \\
    6 datasets & \textbf{0.844} & \textbf{0.285} & \textbf{0.435} \\
    \bottomrule
    \end{tabular}
\end{minipage}%
\hfill
\begin{minipage}[t]{0.45\textwidth}
    \vspace{0pt}
    \centering
    \scriptsize
    \includegraphics[width=\linewidth]{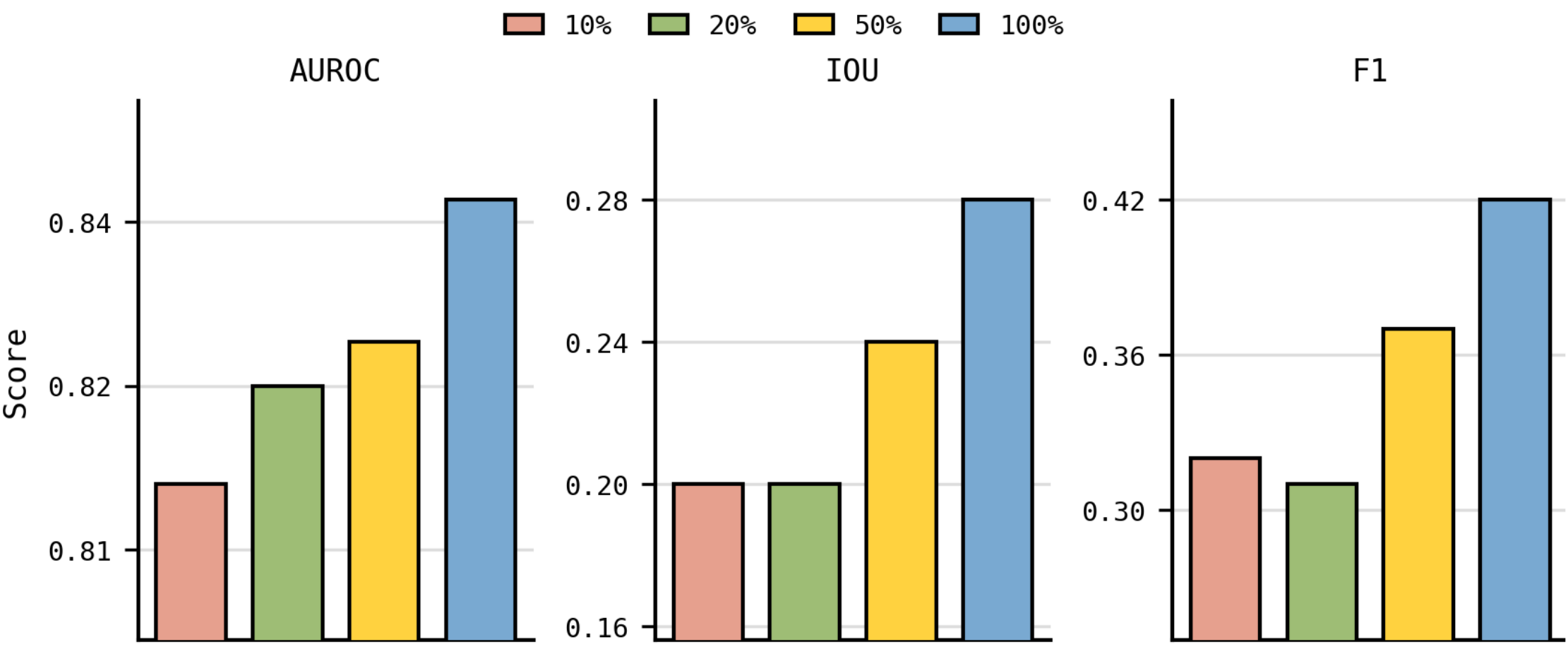}
    \vspace{-2mm}
    \caption{\textbf{Data volume scaling on OOD contact estimation.} Scaling training data volume from 10\% to 100\% improves contact prediction performance on OAKINK2~\cite{zhan2024oakink2}.}
    \label{fig:datascaling}
\end{minipage}
\vspace{-0.6cm}
\end{figure}

\subsection{Effects of Data Scaling}
\label{sec:datascaling}
In this part, we analyze the effects of data scaling from two perspectives: (1) increasing the number of diverse datasets used in training, and (2) scaling the overall data volume by using different proportions of the full training set. 
First, Table~\ref{tab:ablation_datause} shows that leveraging more data sources (from 1 to 6 datasets) steadily improves out-of-domain contact estimation on OAKINK2~\cite{zhan2024oakink2}. 
Second, scaling the total data volume from 10\% to 100\% yields improvements across multiple OOD metrics. Specifically, it enhances contact AUROC, IoU, and F1 on OAKINK2~\cite{zhan2024oakink2} (Figure~\ref{fig:datascaling}), while significantly boosting force-cosine similarity and rise/fall F1 scores on the OpenTouch~\cite{song2025opentouch} benchmark (Figure~\ref{fig:ood_force_vis}(a)). 
These consistent performance gains suggest that EgoTac benefits from both source diversity and data volume, demonstrating the scalability of our formulation and indicating that further improvements can be expected as more tactile-supervised egocentric data becomes available.

\subsection{Understanding Heterogeneous Supervision}
\label{sec:heterogeneous_supervision}
EgoTac combines two types of supervision with complementary physical
meanings: continuous force measurements specify interaction intensity,
while contact annotations provide large-scale information about where and
when interactions occur. To understand the contribution of each supervision
source, we compare three training configurations: (1) EgoTac-SC only,
(2) contact-only HOI datasets only, and (3) their combination. The
contact-only sources include HOT3D~\cite{banerjee2025hot3d}, HOI4D~\cite{liu2022hoi4d}, TACO~\cite{liu2024taco}, H2O~\cite{kwon2021h2o}, and ARCTIC~\cite{fan2023arctic}. Results
are evaluated under the OOD protocols described in
Section~\ref{sec:ood_eval}.

\begin{table}[t]
\centering
\small
\setlength{\tabcolsep}{9pt}
\caption{\textbf{Ablation of supervision sources.}
We compare training on EgoTac-SC, five contact-only HOI datasets, and their
union (Mixed). ``--'' denotes the absence of a force-supervised model.
Tables~\ref{tab:supp_dataset_ablation_oakink}
and~\ref{tab:supp_dataset_ablation_opentouch} report the complete contact
and force-dynamics metrics.}
\label{tab:ablation_dataset}
\vspace{0.2cm}
\begin{tabular}{lcccccc}
\toprule
& \multicolumn{3}{c}{OAKINK2: contact}
& \multicolumn{3}{c}{OpenTouch: force dynamics} \\
\cmidrule(lr){2-4}\cmidrule(lr){5-7}
Training data & AUROC $\uparrow$ & IoU $\uparrow$ & F1 $\uparrow$
& Force-Cosine $\uparrow$
& Rise-F1 $\uparrow$ & Fall-F1 $\uparrow$ \\
\midrule
EgoTac-SC only & 0.60 & 0.07 & 0.12 & 0.745 & \textbf{0.404} & 0.385 \\
Contact-only HOI & \textbf{0.84} & \textbf{0.29} & \textbf{0.44} & -- & -- & -- \\
Mixed & \textbf{0.84} & \textbf{0.29} & \textbf{0.44} & \textbf{0.777} & 0.403 & \textbf{0.403} \\
\bottomrule
\end{tabular}
\end{table}

\paragraph{Complementary roles of force and contact supervision.}
The two supervision sources provide distinct and complementary benefits.
As shown in Table~\ref{tab:ablation_dataset}, on OAKINK2, training with contact-only HOI datasets substantially improves
contact generalization over EgoTac-SC alone, increasing F1 from 0.12 to
0.44. This demonstrates that diverse bare-hand HOI data provide valuable
visual and interaction priors for contact localization. In contrast,
EgoTac-SC provides the only continuous force supervision and is therefore
essential for learning tactile intensity. On OpenTouch, combining EgoTac-SC
with contact-only datasets further improves force cosine from 0.745 to
0.777, indicating that contact supervision can benefit force prediction
through shared visual-temporal representations while measured-force data
remain necessary for recovering continuous tactile dynamics.

\paragraph{Interaction between contact and force supervision.}
The two supervision sources are not independent: contact annotations provide
a structural prior over interaction occurrence, while force annotations
capture the continuous intensity of physical interactions. Through shared
representation learning and the force--contact consistency objective,
contact supervision can improve the generalization of force prediction even
without directly providing force magnitudes. Detailed ablations of this
interaction, including the effect of removing $\mathcal{L}_{\mathrm{cons}}$,
are provided in the Appendix~\ref{sec:supp_more_discussions}.

\section{Conclusion}
\label{sec:conclusion}

We introduce EgoTac, a generalizable framework tailored for predicting dense tactile states directly from in-the-wild egocentric RGB videos. Our contributions are threefold: (1) we construct a massive 5.7M-sample unified dataset by lifting our newly collected real-force EgoTac-SC dataset and eight existing hand-object interaction datasets onto a shared MANO topology; (2) we propose a scalable architecture with a mixed force-contact objective that effectively learns from this heterogeneous supervision; and (3) we demonstrate through comprehensive evaluations that EgoTac achieves accurate in-domain predictions, strong out-of-domain generalization, and impressive zero-shot scaling capabilities on unconstrained real-world scenes. Ultimately, EgoTac provides a foundational step toward extracting rich physical priors from ubiquitous human video corpora.

\clearpage
\section*{Acknowledgements}
This research was conducted with the support of the Shanghai Qi Zhi Institute \& Spirit AI Innovation Program. Funding and support for this work were also provided by the Tsinghua University - Keystone Electrical (Zhejiang) Co.,Ltd Joint Research Center for Embodied Multimodal Artificial Intelligence (JCEMAI). Additionally, we would like to extend our thanks to the Xiongan AI Institute. This project is also supported by Beijing Natural Science Foundation(JR25019).

\bibliography{references}{}

@inproceedings{potamias2025wilor,
  title={Wilor: End-to-end 3d hand localization and reconstruction in-the-wild},
  author={Potamias, Rolandos Alexandros and Zhang, Jinglei and Deng, Jiankang and Zafeiriou, Stefanos},
  booktitle={Proceedings of the Computer Vision and Pattern Recognition Conference},
  pages={12242--12254},
  year={2025}
}

@article{intelligence2025pi_,
  title={pi0.5: a Vision-Language-Action Model with Open-World Generalization},
  author={Intelligence, Physical and Black, Kevin and Brown, Noah and Darpinian, James and Dhabalia, Karan and Driess, Danny and Esmail, Adnan and Equi, Michael and Finn, Chelsea and Fusai, Niccolo and others},
  journal={arXiv preprint arXiv:2504.16054},
  year={2025}
}

@article{bjorck2025gr00t,
  title={Gr00t n1: An open foundation model for generalist humanoid robots},
  author={Bjorck, Johan and Casta{\~n}eda, Fernando and Cherniadev, Nikita and Da, Xingye and Ding, Runyu and Fan, Linxi and Fang, Yu and Fox, Dieter and Hu, Fengyuan and Huang, Spencer and others},
  journal={arXiv preprint arXiv:2503.14734},
  year={2025}
}

@article{zhang2025unitachand,
  title={UniTacHand: Unified Spatio-Tactile Representation for Human to Robotic Hand Skill Transfer},
  author={Zhang, Chi and Cai, Penglin and Yuan, Haoqi and Xu, Chaoyi and Lu, Zongqing},
  journal={arXiv preprint arXiv:2512.21233},
  year={2025}
}

@inproceedings{li2022egocentric,
  title={Egocentric prediction of action target in 3d},
  author={Li, Yiming and Cao, Ziang and Liang, Andrew and Liang, Benjamin and Chen, Luoyao and Zhao, Hang and Feng, Chen},
  booktitle={2022 IEEE/CVF Conference on Computer Vision and Pattern Recognition (CVPR)},
  pages={20971--20980},
  year={2022},
  organization={IEEE}
}

@article{luo2025being,
  title={Being-h0: vision-language-action pretraining from large-scale human videos},
  author={Luo, Hao and Feng, Yicheng and Zhang, Wanpeng and Zheng, Sipeng and Wang, Ye and Yuan, Haoqi and Liu, Jiazheng and Xu, Chaoyi and Jin, Qin and Lu, Zongqing},
  journal={arXiv preprint arXiv:2507.15597},
  year={2025}
}

@inproceedings{liu2024taco,
  title={Taco: Benchmarking generalizable bimanual tool-action-object understanding},
  author={Liu, Yun and Yang, Haolin and Si, Xu and Liu, Ling and Li, Zipeng and Zhang, Yuxiang and Liu, Yebin and Yi, Li},
  booktitle={Proceedings of the IEEE/CVF Conference on Computer Vision and Pattern Recognition},
  pages={21740--21751},
  year={2024}
}

@inproceedings{garcia2018first,
  title={First-person hand action benchmark with rgb-d videos and 3d hand pose annotations},
  author={Garcia-Hernando, Guillermo and Yuan, Shanxin and Baek, Seungryul and Kim, Tae-Kyun},
  booktitle={Proceedings of the IEEE conference on computer vision and pattern recognition},
  pages={409--419},
  year={2018}
}

@article{oquab2023dinov2,
  title={Dinov2: Learning robust visual features without supervision},
  author={Oquab, Maxime and Darcet, Timoth{\'e}e and Moutakanni, Th{\'e}o and Vo, Huy and Szafraniec, Marc and Khalidov, Vasil and Fernandez, Pierre and Haziza, Daniel and Massa, Francisco and El-Nouby, Alaaeldin and others},
  journal={arXiv preprint arXiv:2304.07193},
  year={2023}
}

@article{kareer2025emergence,
  title={Emergence of Human to Robot Transfer in Vision-Language-Action Models},
  author={Kareer, Simar and Pertsch, Karl and Darpinian, James and Hoffman, Judy and Xu, Danfei and Levine, Sergey and Finn, Chelsea and Nair, Suraj},
  journal={arXiv preprint arXiv:2512.22414},
  year={2025}
}

@article{yuan2025motiontrans,
  title={Motiontrans: Human vr data enable motion-level learning for robotic manipulation policies},
  author={Yuan, Chengbo and Zhou, Rui and Liu, Mengzhen and Hu, Yingdong and Wang, Shengjie and Yi, Li and Wen, Chuan and Zhang, Shanghang and Gao, Yang},
  journal={arXiv preprint arXiv:2509.17759},
  year={2025}
}

@article{zheng2026egoscale,
  title={Egoscale: Scaling dexterous manipulation with diverse egocentric human data},
  author={Zheng, Ruijie and Niu, Dantong and Xie, Yuqi and Wang, Jing and Xu, Mengda and Jiang, Yunfan and Casta{\~n}eda, Fernando and Hu, Fengyuan and Tan, You Liang and Fu, Letian and others},
  journal={arXiv preprint arXiv:2602.16710},
  year={2026}
}

@inproceedings{kareer2025egomimic,
  title={Egomimic: Scaling imitation learning via egocentric video},
  author={Kareer, Simar and Patel, Dhruv and Punamiya, Ryan and Mathur, Pranay and Cheng, Shuo and Wang, Chen and Hoffman, Judy and Xu, Danfei},
  booktitle={2025 IEEE International Conference on Robotics and Automation (ICRA)},
  pages={13226--13233},
  year={2025},
  organization={IEEE}
}

@article{li2026egocentric,
  title={Egocentric World Model for Photorealistic Hand-Object Interaction Synthesis},
  author={Li, Dayou and Liu, Lulin and Liu, Bangya and Zhou, Shijie and Feng, Jiu and Lu, Ziqi and Zheng, Minghui and You, Chenyu and Fan, Zhiwen},
  journal={arXiv preprint arXiv:2603.13615},
  year={2026}
}

@inproceedings{fan2023arctic,
  title={ARCTIC: A dataset for dexterous bimanual hand-object manipulation},
  author={Fan, Zicong and Taheri, Omid and Tzionas, Dimitrios and Kocabas, Muhammed and Kaufmann, Manuel and Black, Michael J and Hilliges, Otmar},
  booktitle={Proceedings of the IEEE/CVF conference on computer vision and pattern recognition},
  pages={12943--12954},
  year={2023}
}

@inproceedings{liu2022hoi4d,
  title={Hoi4d: A 4d egocentric dataset for category-level human-object interaction},
  author={Liu, Yunze and Liu, Yun and Jiang, Che and Lyu, Kangbo and Wan, Weikang and Shen, Hao and Liang, Boqiang and Fu, Zhoujie and Wang, He and Yi, Li},
  booktitle={Proceedings of the IEEE/CVF Conference on Computer Vision and Pattern Recognition},
  pages={21013--21022},
  year={2022}
}

@article{delpreto2022actionsense,
  title={ActionSense: A multimodal dataset and recording framework for human activities using wearable sensors in a kitchen environment},
  author={DelPreto, Joseph and Liu, Chao and Luo, Yiyue and Foshey, Michael and Li, Yunzhu and Torralba, Antonio and Matusik, Wojciech and Rus, Daniela},
  journal={Advances in Neural Information Processing Systems},
  volume={35},
  pages={13800--13813},
  year={2022}
}

@inproceedings{grauman2022ego4d,
  title={Ego4d: Around the world in 3,000 hours of egocentric video},
  author={Grauman, Kristen and Westbury, Andrew and Byrne, Eugene and Chavis, Zachary and Furnari, Antonino and Girdhar, Rohit and Hamburger, Jackson and Jiang, Hao and Liu, Miao and Liu, Xingyu and others},
  booktitle={Proceedings of the IEEE/CVF conference on computer vision and pattern recognition},
  pages={18995--19012},
  year={2022}
}

@article{hoque2025egodex,
  title={Egodex: Learning dexterous manipulation from large-scale egocentric video},
  author={Hoque, Ryan and Huang, Peide and Yoon, David J and Sivapurapu, Mouli and Zhang, Jian},
  journal={arXiv preprint arXiv:2505.11709},
  year={2025}
}

@inproceedings{damen2018scaling,
  title={Scaling egocentric vision: The epic-kitchens dataset},
  author={Damen, Dima and Doughty, Hazel and Farinella, Giovanni Maria and Fidler, Sanja and Furnari, Antonino and Kazakos, Evangelos and Moltisanti, Davide and Munro, Jonathan and Perrett, Toby and Price, Will and others},
  booktitle={Proceedings of the European conference on computer vision (ECCV)},
  pages={720--736},
  year={2018}
}

@inproceedings{banerjee2025hot3d,
  title={Hot3d: Hand and object tracking in 3d from egocentric multi-view videos},
  author={Banerjee, Prithviraj and Shkodrani, Sindi and Moulon, Pierre and Hampali, Shreyas and Han, Shangchen and Zhang, Fan and Zhang, Linguang and Fountain, Jade and Miller, Edward and Basol, Selen and others},
  booktitle={Proceedings of the IEEE/CVF Conference on Computer Vision and Pattern Recognition},
  pages={7061--7071},
  year={2025}
}

@article{wang2024ho,
  title={Ho-cap: A capture system and dataset for 3d reconstruction and pose tracking of hand-object interaction},
  author={Wang, Jikai and Zhang, Qifan and Chao, Yu-Wei and Wen, Bowen and Guo, Xiaohu and Xiang, Yu},
  journal={arXiv preprint arXiv:2406.06843},
  year={2024}
}

@inproceedings{kwon2021h2o,
  title={H2o: Two hands manipulating objects for first person interaction recognition},
  author={Kwon, Taein and Tekin, Bugra and St{\"u}hmer, Jan and Bogo, Federica and Pollefeys, Marc},
  booktitle={Proceedings of the IEEE/CVF international conference on computer vision},
  pages={10138--10148},
  year={2021}
}

@inproceedings{zhan2024oakink2,
  title={Oakink2: A dataset of bimanual hands-object manipulation in complex task completion},
  author={Zhan, Xinyu and Yang, Lixin and Zhao, Yifei and Mao, Kangrui and Xu, Hanlin and Lin, Zenan and Li, Kailin and Lu, Cewu},
  booktitle={Proceedings of the IEEE/CVF Conference on Computer Vision and Pattern Recognition},
  pages={445--456},
  year={2024}
}

@article{dessalene2026feel,
  title={FEEL (Force-Enhanced Egocentric Learning): A Dataset for Physical Action Understanding},
  author={Dessalene, Eadom and He, Botao and Maynord, Michael and Tussa, Yonatan and Mantripragada, Pavan and Karabatis, Yianni and Roy, Nirupam and Aloimonos, Yiannis},
  journal={arXiv preprint arXiv:2603.15847},
  year={2026}
}

@inproceedings{jung2025learning,
  title={Learning Dense Hand Contact Estimation from Imbalanced Data},
  author={Jung, Daniel Sungho and Lee, Kyoung Mu},
  booktitle={The Thirty-ninth Annual Conference on Neural Information Processing Systems},
  year={2025}
}

@article{yuan2017gelsight,
  title={Gelsight: High-resolution robot tactile sensors for estimating geometry and force},
  author={Yuan, Wenzhen and Dong, Siyuan and Adelson, Edward H},
  journal={Sensors},
  volume={17},
  number={12},
  pages={2762},
  year={2017},
  publisher={MDPI}
}

@inproceedings{ehsani2020use,
  title={Use the force, luke! learning to predict physical forces by simulating effects},
  author={Ehsani, Kiana and Tulsiani, Shubham and Gupta, Saurabh and Farhadi, Ali and Gupta, Abhinav},
  booktitle={Proceedings of the IEEE/CVF Conference on Computer Vision and Pattern Recognition},
  pages={224--233},
  year={2020}
}

@article{pham2017hand,
  title={Hand-object contact force estimation from markerless visual tracking},
  author={Pham, Tu-Hoa and Kyriazis, Nikolaos and Argyros, Antonis A and Kheddar, Abderrahmane},
  journal={IEEE transactions on pattern analysis and machine intelligence},
  volume={40},
  number={12},
  pages={2883--2896},
  year={2017},
  publisher={IEEE}
}

@article{fallahinia2022real,
  title={Real-time tactile grasp force sensing using fingernail imaging via deep neural networks},
  author={Fallahinia, Navid and Mascaro, Stephen A},
  journal={IEEE Robotics and Automation Letters},
  volume={7},
  number={3},
  pages={6558--6565},
  year={2022},
  publisher={IEEE}
}

@inproceedings{pham2015towards,
  title={Towards force sensing from vision: Observing hand-object interactions to infer manipulation forces},
  author={Pham, Tu-Hoa and Kheddar, Abderrahmane and Qammaz, Ammar and Argyros, Antonis A},
  booktitle={Proceedings of the IEEE conference on computer vision and pattern recognition},
  pages={2810--2819},
  year={2015}
}

@inproceedings{grady2024pressurevision++,
  title={Pressurevision++: Estimating fingertip pressure from diverse rgb images},
  author={Grady, Patrick and Collins, Jeremy A and Tang, Chengcheng and Twigg, Christopher D and Aneja, Kunal and Hays, James and Kemp, Charles C},
  booktitle={Proceedings of the IEEE/CVF Winter Conference on Applications of Computer Vision},
  pages={8698--8708},
  year={2024}
}

@inproceedings{grady2022pressurevision,
  title={PressureVision: estimating hand pressure from a single RGB image},
  author={Grady, Patrick and Tang, Chengcheng and Brahmbhatt, Samarth and Twigg, Christopher D and Wan, Chengde and Hays, James and Kemp, Charles C},
  booktitle={European Conference on Computer Vision},
  pages={328--345},
  year={2022},
  organization={Springer}
}

@inproceedings{zhao2025egopressure,
  title={Egopressure: A dataset for hand pressure and pose estimation in egocentric vision},
  author={Zhao, Yiming and Kwon, Taein and Streli, Paul and Pollefeys, Marc and Holz, Christian},
  booktitle={Proceedings of the Computer Vision and Pattern Recognition Conference},
  pages={27727--27738},
  year={2025}
}

@article{chen2020estimating,
  title={Estimating fingertip forces, torques, and local curvatures from fingernail images},
  author={Chen, Nutan and Westling, G{\"o}ran and Edin, Benoni B and Van Der Smagt, Patrick},
  journal={Robotica},
  volume={38},
  number={7},
  pages={1242--1262},
  year={2020},
  publisher={Cambridge University Press}
}

@inproceedings{hanai2023force,
  title={Force map: Learning to predict contact force distribution from vision},
  author={Hanai, Ryo and Domae, Yukiyasu and Ramirez-Alpizar, Ixchel G and Leme, Bruno and Ogata, Tetsuya},
  booktitle={2023 IEEE/RSJ International Conference on Intelligent Robots and Systems (IROS)},
  pages={3129--3136},
  year={2023},
  organization={IEEE}
}

@article{song2025opentouch,
  title={OPENTOUCH: Bringing Full-Hand Touch to Real-World Interaction},
  author={Song, Yuxin Ray and Li, Jinzhou and Fu, Rao and Murphy, Devin and Zhou, Kaichen and Shiv, Rishi and Li, Yaqi and Xiong, Haoyu and Owens, Crystal Elaine and Du, Yilun and others},
  journal={arXiv preprint arXiv:2512.16842},
  year={2025}
}

@inproceedings{li2019connecting,
  title={Connecting touch and vision via cross-modal prediction},
  author={Li, Yunzhu and Zhu, Jun-Yan and Tedrake, Russ and Torralba, Antonio},
  booktitle={Proceedings of the IEEE/CVF Conference on Computer Vision and Pattern Recognition},
  pages={10609--10618},
  year={2019}
}

@inproceedings{li2023learning,
  title={Learning to jointly understand visual and tactile signals},
  author={Li, Yichen and Du, Yilun and Liu, Chao and Williams, Francis and Foshey, Michael and Eckart, Benjamin and Kautz, Jan and Tenenbaum, Joshua B and Torralba, Antonio and Matusik, Wojciech},
  booktitle={The Twelfth International Conference on Learning Representations},
  year={2023}
}

@inproceedings{gao2023objectfolder,
  title={The objectfolder benchmark: Multisensory learning with neural and real objects},
  author={Gao, Ruohan and Dou, Yiming and Li, Hao and Agarwal, Tanmay and Bohg, Jeannette and Li, Yunzhu and Fei-Fei, Li and Wu, Jiajun},
  booktitle={Proceedings of the IEEE/CVF Conference on Computer Vision and Pattern Recognition},
  pages={17276--17286},
  year={2023}
}

@inproceedings{huang2022capturing,
  title={Capturing and inferring dense full-body human-scene contact},
  author={Huang, Chun-Hao P and Yi, Hongwei and H{\"o}schle, Markus and Safroshkin, Matvey and Alexiadis, Tsvetelina and Polikovsky, Senya and Scharstein, Daniel and Black, Michael J},
  booktitle={CVPR},
  year={2022}
}

@inproceedings{tripathi2023deco,
  title={{DECO}: Dense estimation of {3D} human-scene contact in the wild},
  author={Tripathi, Shashank and Chatterjee, Agniv and Passy, Jean-Claude and Yi, Hongwei and Tzionas, Dimitrios and Black, Michael J},
  booktitle={ICCV},
  year={2023}
}

@inproceedings{grady2021contactopt,
  title={{ContactOpt}: Optimizing contact to improve grasps},
  author={Grady, Patrick and Tang, Chengcheng and Twigg, Christopher D and Vo, Minh and Brahmbhatt, Samarth and Kemp, Charles C},
  booktitle={CVPR},
  year={2021}
}

@article{romero2017embodied,
  title={Embodied hands: Modeling and capturing hands and bodies together},
  author={Romero, Javier and Tzionas, Dimitrios and Black, Michael J},
  journal={ACM TOG},
  year={2017}
}

@inproceedings{liu2024easyhoi,
  title={{EasyHOI}: Unleashing the Power of Large Models for Reconstructing Hand-Object Interactions in the Wild},
  author={Liu, Yumeng and Long, Xiaoxiao and Yang, Zemin and Liu, Yuan and Habermann, Marc and Theobalt, Christian and Ma, Yuexin and Wang, Wenping},
  booktitle={CVPR},
  year={2025}
}

@inproceedings{brahmbhatt2019contactdb,
  title={{ContactDB}: Analyzing and predicting grasp contact via thermal imaging},
  author={Brahmbhatt, Samarth and Ham, Cusuh and Kemp, Charles C and Hays, James},
  booktitle={CVPR},
  year={2019}
}

@inproceedings{loshchilov2018decoupled,
    title={Decoupled Weight Decay Regularization},
    author={Loshchilov, Ilya and Hutter, Frank},
    booktitle={ICLR},
    year={2019}
}

@article{zhang2026glove2hand,
  title={Glove2Hand: Synthesizing Natural Hand-Object Interaction from Multi-Modal Sensing Gloves},
  author={Zhang, Xinyu and Kou, Ziyi and Qin, Chuan and Huang, Mia and Ristani, Ergys and Kumar, Ankit and Chen, Lele and He, Kun and Boularias, Abdeslam and Guan, Li},
  journal={arXiv preprint arXiv:2603.20850},
  year={2026}
}
\bibliographystyle{plain}

\clearpage
\appendix
\section{Appendix Overview}\label{sec:appendix}

\setcounter{table}{0}
\setcounter{figure}{0}
\renewcommand{\thetable}{X\arabic{table}}
\renewcommand{\thefigure}{X\arabic{figure}}

This supplementary material provides additional details on dataset construction, model implementation, and evaluation for \emph{EgoTac}.
It is organized as follows:
\begin{itemize}
\item \ref{sec:supp_egotac_sc}. Details of the self-collected dataset (EgoTac-SC)
\item \ref{sec:supp_other_datasets}. Details of other datasets
\item \ref{sec:supp_impl}. Details of EgoTac implementation
\item \ref{sec:supp_metrics}. Details of metrics
\item \ref{sec:supp_more_results}. More experimental results
\item \ref{sec:supp_more_discussions}. More discussions
\item \ref{sec:supp_limitations}. Limitations and societal impacts
\end{itemize}

\section{Details of the self-collected dataset (EgoTac-SC)}
\label{sec:supp_egotac_sc}

This section provides additional details on the construction, calibration, post-processing, and organization of our self-collected \emph{EgoTac-SC} dataset. EgoTac-SC is collected with a wearable multimodal platform that synchronizes egocentric visual observations, hand-pose tracking, and full-hand tactile measurements. We convert the raw streams into the unified MANO-aligned representation used throughout the paper, enabling joint training with both force-supervised and contact-only datasets.

\subsection{Dataset collection}
\label{sec:supp_egotac_sc_collection}

\begin{wrapfigure}{t}{0.5\textwidth}
\begin{center}
\includegraphics[width=0.93\linewidth]{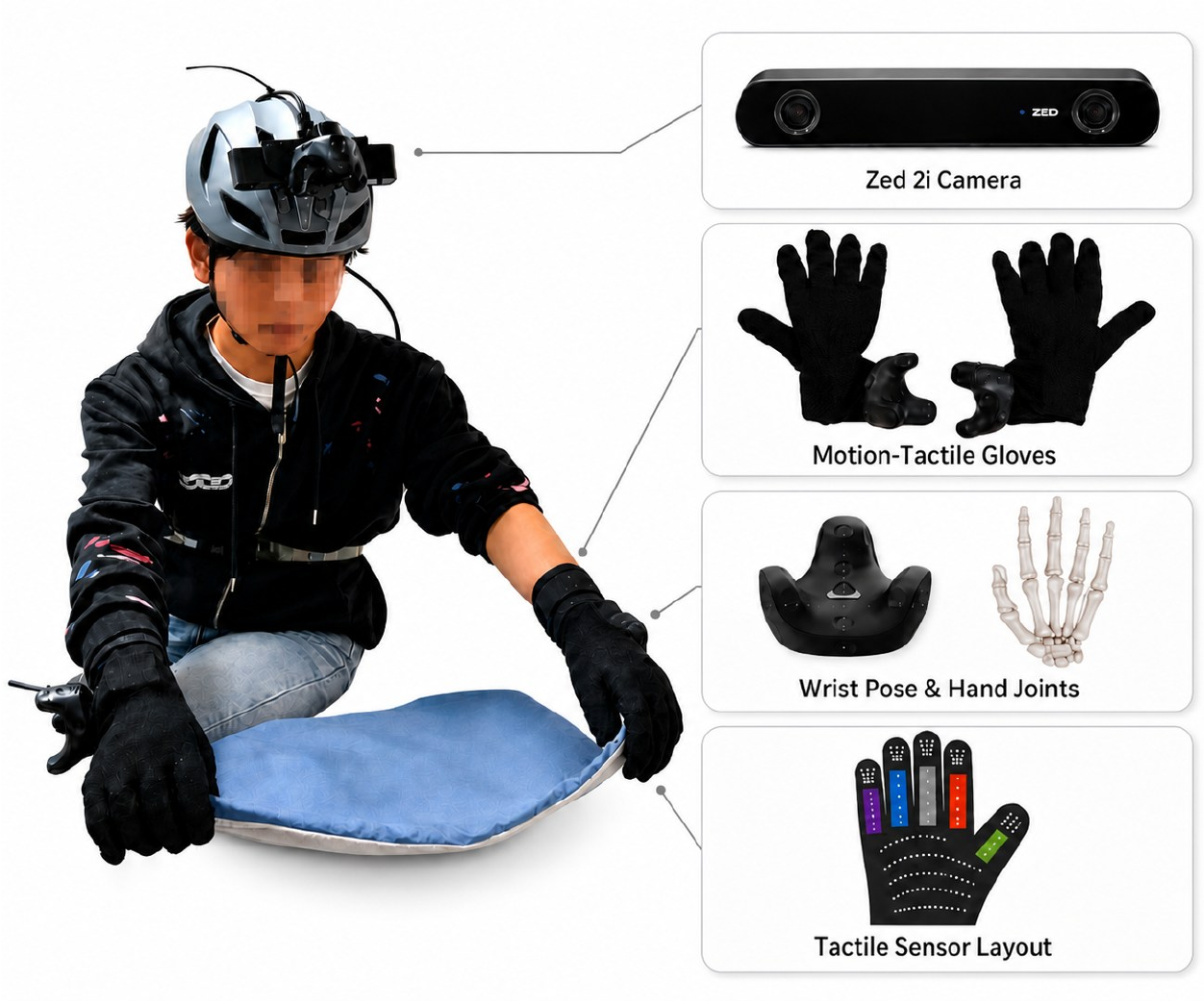}
\end{center}
\caption{
\textbf{Data-collection hardware.} Our capture platform includes three synchronized components: an egocentric RGB-D camera, wearable motion-tactile gloves, and wrist-pose trackers.}
\label{fig:supp_hardware}
\vspace{-0.4cm}
\end{wrapfigure}

\paragraph{Hardware.}
\label{sec:supp_egotac_sc_hardware}

Figure~\ref{fig:supp_hardware} illustrates the capture hardware used for our data collection. The platform contains three synchronized sensing components: an egocentric RGB-D camera, wearable motion-tactile gloves, and wrist-pose tracking devices. The visual stream is captured by a head-mounted ZED 2i camera at 1080p resolution, with optional point-cloud export. For hand tracking and tactile sensing, each hand is equipped with an HTC VIVE Tracker and an in-house motion-tactile glove. Each glove includes 11 high-precision nine-axis inertial modules and a dense force-sensing matrix with more than 120 sensing points. The inertial modules and VIVE Tracker provide hand-pose-related signals (20 hand joint poses plus one wrist pose), while the force matrix records distributed tactile measurements over fingers and palm. Together, these sensors capture synchronized egocentric vision, hand motion, and full-hand tactile interaction during natural manipulation.

\paragraph{Collection procedure.}
\label{sec:supp_egotac_sc_collection_procedure}

Before recording, the head-mounted camera and wearable hand devices are connected to the acquisition terminal. The collection process follows a fixed protocol: we first launch the transmission and acquisition software, then perform pose calibration for the wearable trackers. After calibration, participants perform assigned manipulation tasks while the system records synchronized visual, pose, and tactile streams. Each sequence is terminated through the acquisition interface, after which the raw streams are saved for offline synchronization and conversion.

\paragraph{Raw tactile channel organization.}

Each hand contains a dense force-sensing matrix. In the raw byte layout, finger and palm force channels are interleaved with several bend-related channels. We exclude bend channels from supervision and retain only force channels. Specifically, each hand uses 60 finger-force channels and 72 palm-force channels, resulting in a 132-dimensional tactile vector \(s \in \mathbb{R}^{132}\). The five bend-related channels are not used for tactile-force annotation.

\subsection{Tactile post-processing and MANO hand mapping}
\label{sec:supp_egotac_sc_postprocess}

\paragraph{Temporal alignment and interpolation.}

Raw streams are stored with individual timestamps. We parse wrist, hand-pose, and tactile streams and resample them to a unified timeline. For pose signals involving rigid transformations or rotations, we use interpolation suitable for SE(3)-type signals. For tactile force readings, we use linear interpolation. This yields frame-level temporal alignment across visual, pose, and tactile observations.

\paragraph{RGB decoding and synchronization.}

Egocentric video frames are decoded from the recorded camera stream and synchronized to the unified timeline. The aligned RGB frames are then resized and cropped to the training resolution.

\paragraph{Coordinate normalization.}

Because the camera, wrist trackers, and glove sensors use different coordinate conventions, we apply a fixed geometric transformation chain. This step normalizes camera, tracker, and hand-pose signals into a consistent convention before mapping tactile values to the MANO surface.

\begin{figure*}[t]
\begin{center}
\includegraphics[width=0.7\linewidth]{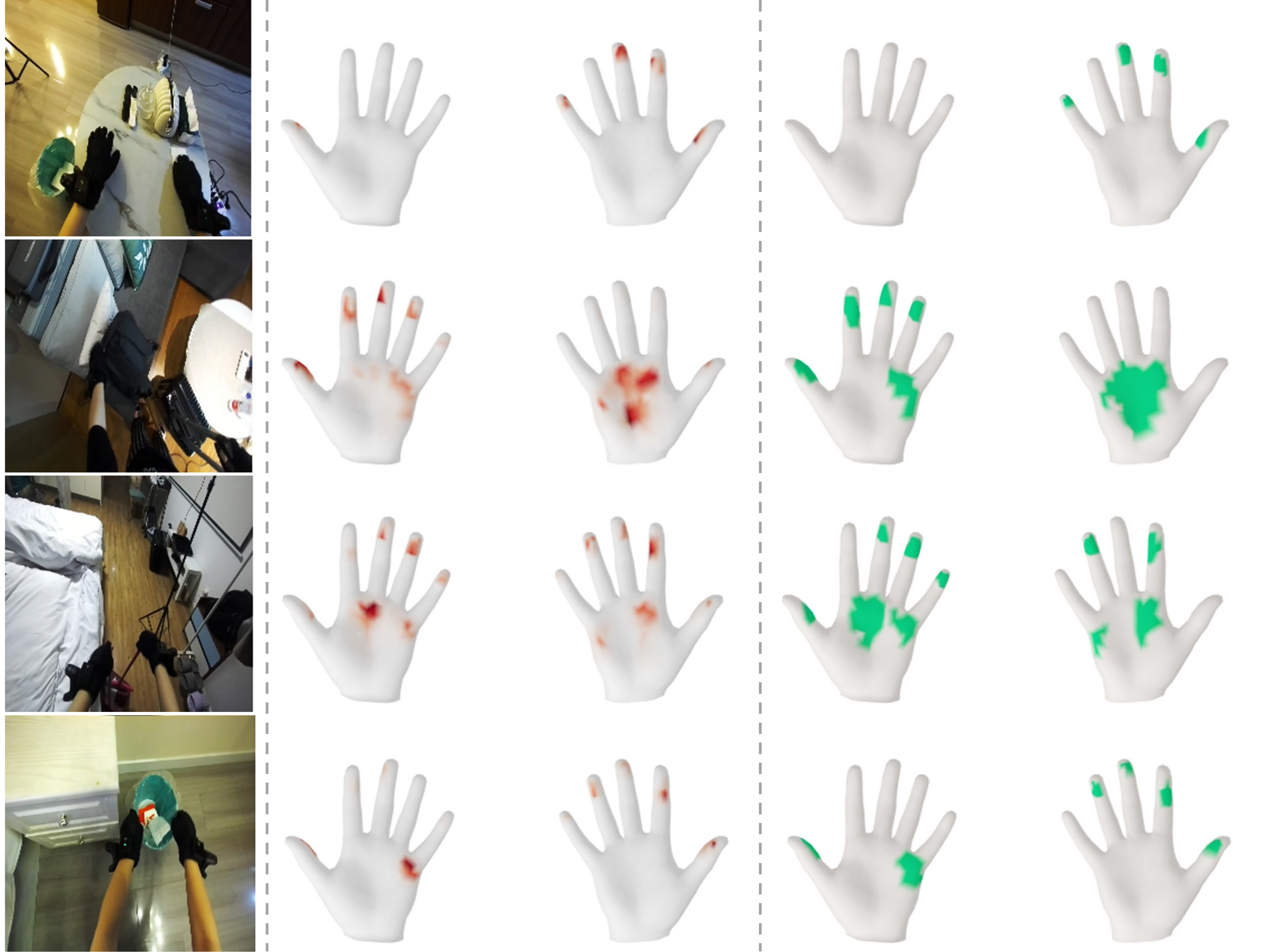}
\end{center}
\vspace{-0.2cm}
\caption{
\textbf{Representative samples from EgoTac-SC.} Each example shows egocentric RGB together with MANO-aligned tactile-force maps and binary contact labels.}
\label{fig:supp_vis_data_egotac}
\vspace{-0.2cm}
\end{figure*}

\paragraph{Tactile-to-MANO mapping.}

After temporal alignment and coordinate normalization, we map the per-hand tactile vector to the MANO mesh. Each MANO hand contains \(V=778\) vertices. Following \cite{zhang2025unitachand}, we distribute sparse glove-force readings to dense MANO vertices through predefined region patches and bilinear sensor weights.

Let \(s\in\mathbb{R}^{132}\) denote the force vector of one hand at a single frame. For each MANO vertex \(v\), let \(\mathcal{W}(v)\) be the set of neighboring tactile sensors used to interpolate the force value at that vertex. The mapped force value is computed as
\begin{equation}
F_v = \sum_{k\in\mathcal{W}(v)} w_{v,k}\, s_k,
\qquad
\sum_{k\in\mathcal{W}(v)} w_{v,k}=1,
\end{equation}
where \(w_{v,k}\) is the interpolation weight between MANO vertex \(v\) and tactile sensor \(k\). This produces a dense tactile field $F \in \mathbb{R}^{V}$ for each hand. Binary contact labels are then obtained by thresholding the mapped tactile force:
\begin{equation}
C_v = \mathbbm{1}[F_v > \tau_c],
\end{equation}
where \(\tau_c\) is set to 0.05 N during preprocessing. After mapping both hands, each frame contains dense tactile and contact annotations over \(2V=1556\) MANO vertices:
\begin{equation}
F \in \mathbb{R}^{2V}, \qquad C \in \{0,1\}^{2V}.
\end{equation}
This representation is consistent with the unified data format described in Sec.~\ref{sec:dataset_format} and allows EgoTac-SC to be trained jointly with contact-only datasets.

\subsection{Visualization of samples}
\label{sec:supp_egotac_sc_vis}

Representative EgoTac-SC samples are shown in Figure~\ref{fig:supp_vis_data_egotac}. They illustrate diverse contact locations and force magnitudes under natural manipulation. Force annotations are visualized as continuous heatmaps on the MANO surface, while contact annotations are shown as binary active regions obtained by thresholding tactile force.

\begin{figure*}[t]
\begin{center}
\includegraphics[width=0.75\linewidth]{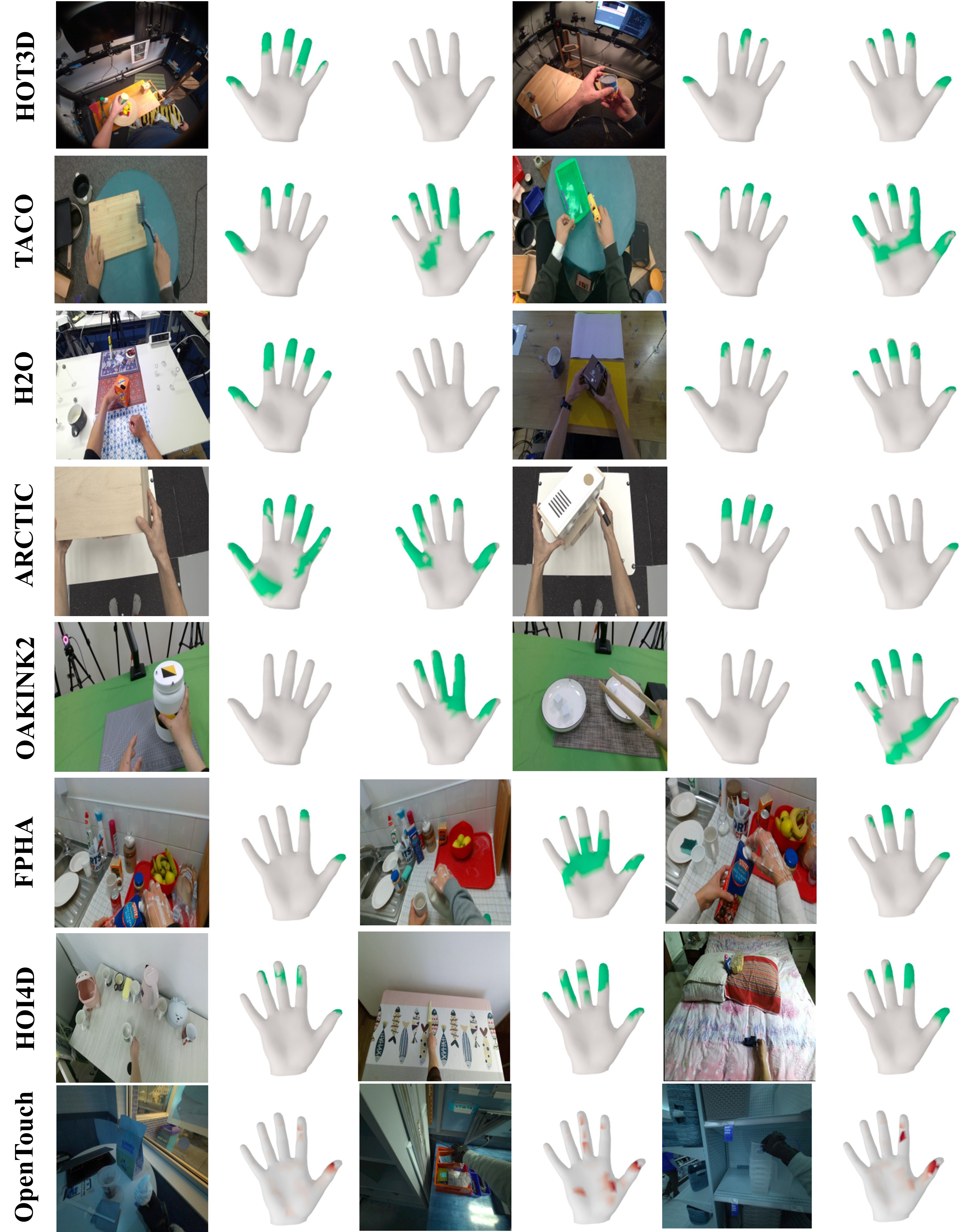}
\end{center}
\caption{
\textbf{Examples from eight auxiliary HOI datasets.} Contact or tactile annotations are converted to a shared MANO topology to enable joint training across datasets.}
\label{fig:supp_vis_data_other}
\end{figure*}

\section{Details of other datasets}
\label{sec:supp_other_datasets}

This section describes how auxiliary datasets are normalized into a unified supervision protocol so that heterogeneous sources can be trained jointly without changing the model interface.

\subsection{MANO hand contact label deriving}
\label{sec:supp_other_contact_deriving}

For datasets without native dense force labels (HOT3D~\cite{banerjee2025hot3d}, TACO~\cite{liu2024taco}, HOI4D~\cite{liu2022hoi4d}, H2O~\cite{kwon2021h2o}, ARCTIC~\cite{fan2023arctic}, OAKINK2~\cite{zhan2024oakink2}, and FPHA~\cite{garcia2018first}), we derive MANO-vertex contact labels from geometric proximity between reconstructed hand and object meshes. For frame \(t\), vertex \(v\), and object vertex set \(\mathcal{O}_t\), we use:
\begin{equation}
C_{t,v}=\mathbbm{1}\!\left[\min_{u\in\mathcal{O}_t}\left\|\mathbf{x}^{\mathrm{hand}}_{t,v}-\mathbf{x}^{\mathrm{obj}}_{t,u}\right\|_2\le \delta\right],
\end{equation}
where \(\delta\) is a fixed contact threshold set to 1 cm following \cite{jung2025learning}. This produces binary contact targets on the same MANO topology as our force supervision.

\paragraph{Unified conversion protocol.}
Although source datasets differ in annotation style and storage format, we apply a consistent conversion pipeline:
\begin{itemize}
\item Decode egocentric RGB frames and timestamps.
\item Reconstruct hand meshes in a canonical MANO space.
\item Reconstruct or load object geometry and align it to frame coordinates.
\item Compute contact labels using nearest-distance thresholding.
\item Export a standardized episode-level format with RGB, MANO contact fields, timestamps, and episode boundaries.
\end{itemize}

\subsection{OpenTouch dataset process}
\label{sec:supp_other_opentouch}

For OpenTouch~\cite{song2025opentouch}, which provides real tactile measurements with a different sensor layout, we follow the official processing instructions to convert raw tactile grids into MANO-aligned tactile vectors. The conversion includes force normalization, spatial alignment from sensor coordinates to MANO vertices, and temporal synchronization with egocentric RGB.

\subsection{Visualization of samples}
\label{sec:supp_other_vis}

Examples from external datasets are shown in Figure~\ref{fig:supp_vis_data_other}, highlighting diversity in interaction type, contact sparsity, hand-viewpoint shifts, and scene composition.

\section{Details of EgoTac implementation}
\label{sec:supp_impl}

This section summarizes the implementation choices used in our main experiments, including data processing, optimization, and objective design.

\subsection{Data augmentation}
\label{sec:supp_impl_aug}

We apply temporally consistent RGB augmentation within each observation window:
\begin{itemize}
\item Color jitter and resized random crop.
\item View-scale augmentation with scale range \([0.9, 1.1]\), applied with probability \(0.8\).
\item Left-right mirror augmentation with probability \(0.5\), applied to the selected asymmetric data source HOI4D~\cite{liu2022hoi4d}.
\end{itemize}

\subsection{Dataset sampling}
\label{sec:supp_impl_sampling}

We use mixed-source temperature sampling to avoid source starvation during multi-dataset training. For source \(i\) with pool size \(n_i\), the sampling weight is
\begin{equation}
w_i = n_i^{\tau}, \qquad p_i = \frac{w_i}{\sum_j w_j},
\end{equation}
where \(\tau=0.5\) in our default setup. We additionally enforce per-source minimum quotas to prevent under-training of small datasets.
Let \(E\) denote epoch size (default: full active training pool size). The initial target sample count of source \(i\) is
\begin{equation}
\tilde{m}_i = p_i E.
\end{equation}
With minimum quota \(q_i\in[0,1]\), the lower-bound count is
\begin{equation}
m_i^{\min} = \lfloor q_i E \rfloor,
\end{equation}
and the final per-source target \(m_i\) satisfies
\begin{equation}
m_i \ge m_i^{\min}, \qquad \sum_i m_i = E.
\end{equation}
In practice, we use conservative non-zero quotas for low-resource sources to stabilize optimization and preserve source diversity.

\begin{table}[t]
\centering
\small
\caption{Main training hyperparameters used in the primary setting.}
\label{tab:supp_hparams}
\begin{tabular}{ll}
\hline
Item & Value \\
\hline
Observation horizon \(T\) & 4 \\
Tactile horizon \(K\) & 1 \\
Input RGB size & \(224\times224\) \\
Output dimension & 1556 (two-hand MANO vertices) \\
Batch size (per step) & 32 $\times$ 4 \\
Gradient accumulation & 1 \\
Optimizer & AdamW \\
Learning rate & \(1\times10^{-4}\) \\
Final learning rate & \(1\times10^{-5}\) \\
Weight decay & \(1\times10^{-5}\) \\
LR scheduler & cosine \\
Warm-up ratio & 0.05 \\
Max grad norm & 5.0 \\
Mixed precision & bf16 \\
Decoder layers / heads / width & 12 / 12 / 768 \\
Vision encoder & DINOv2-Base~\cite{oquab2023dinov2} \\
Temperature sampling \(\tau\) & 0.5 \\
Force loss weight \(\lambda_F\) & 1.0 \\
Contact loss weight \(\lambda_C\) & 1.0 \\
Active-region loss weight \(\lambda_A\) & 0.2 \\
Consistency loss weight \(\lambda_{\mathrm{cons}}\) & 0.05 \\
\hline
\end{tabular}
\vspace{-0.2cm}
\end{table}

\subsection{Training objective}
\label{sec:supp_impl_loss}

Let \(\hat{\mathbf{F}}, \mathbf{F}\in\mathbb{R}^{K\times 2V}\) denote predicted and target force, and \(\hat{\mathbf{Z}}, \mathbf{C}\in\mathbb{R}^{K\times 2V}\) denote predicted contact logits and target contact labels. Let \(\mathbf{M}^F,\mathbf{M}^C\in\{0,1\}^{K\times 2V}\) be force/contact valid masks.
We define masked mean as
\begin{equation}
\langle \mathbf{A}\rangle_{\mathbf{M}}=
\frac{\sum_{t,v} A_{t,v}M_{t,v}}{\sum_{t,v} M_{t,v}+\epsilon}.
\end{equation}

Force loss (MSE) is
\begin{equation}
\mathcal{L}_F
=
\left\langle (\hat{\mathbf{F}}-\mathbf{F})^2 \right\rangle_{\mathbf{M}^F}.
\end{equation}

Active-region loss uses active mask \(\mathbf{M}^{A}\):
\begin{equation}
\mathbf{M}^{A}_{t,v}
=
\mathbbm{1}\!\left[\mathbf{F}^{\mathrm{raw}}_{t,v}>\delta\right]\mathbf{M}^{F}_{t,v},
\end{equation}
where \(\delta=0.05\) in our default setting, and
\begin{equation}
\mathcal{L}_A
=
\left\langle \mathrm{SmoothL1}(\hat{\mathbf{F}},\mathbf{F}) \right\rangle_{\mathbf{M}^{A}}.
\end{equation}

Contact loss with positive reweighting \(\alpha\) is
\begin{equation}
\mathcal{L}_C
=
\left\langle
\mathrm{BCEWithLogits}\!\left(\hat{\mathbf{Z}},\mathbf{C}; \alpha\right)
\right\rangle_{\mathbf{M}^C},
\end{equation}
where \(\alpha=2.0\) in our default setting.

For consistency, we derive force-branch contact logits
\begin{equation}
\hat{\mathbf{Z}}^{F}
=
\frac{\hat{\mathbf{F}}^{\mathrm{raw}}-\theta}{\tau_c},
\end{equation}
with \(\theta=0.05\) and \(\tau_c=0.03\), and use detached contact-head probabilities as soft targets:
\begin{equation}
\mathcal{L}_{\mathrm{cons}}
=
\left\langle
\mathrm{BCEWithLogits}\!\left(
\hat{\mathbf{Z}}^{F},
\sigma(\hat{\mathbf{Z}})^{\mathrm{stopgrad}}
\right)
\right\rangle_{\mathbf{M}^{C}}.
\end{equation}

The total objective is
\begin{equation}
\mathcal{L}
=
\lambda_F\mathcal{L}_F
+\lambda_A\mathcal{L}_A
+\lambda_C\mathcal{L}_C
+\lambda_{\mathrm{cons}}\mathcal{L}_{\mathrm{cons}},
\end{equation}
with \((\lambda_F,\lambda_A,\lambda_C,\lambda_{\mathrm{cons}})=(1.0,0.2,1.0,0.05)\) for the main setting.

\subsection{Hyperparameter settings}
\label{sec:supp_impl_hparams}

Our model uses a pretrained ViT visual encoder and a 12-layer AdaLN Transformer decoder (\(d{=}768\), 12 heads). Training uses AdamW~\cite{loshchilov2018decoupled} with cosine learning-rate decay and warm-up, with mixed-precision distributed optimization and gradient clipping.
Key hyperparameters are listed in Table~\ref{tab:supp_hparams}.

\begin{figure*}[t]
\begin{center}
\includegraphics[width=0.8\linewidth]{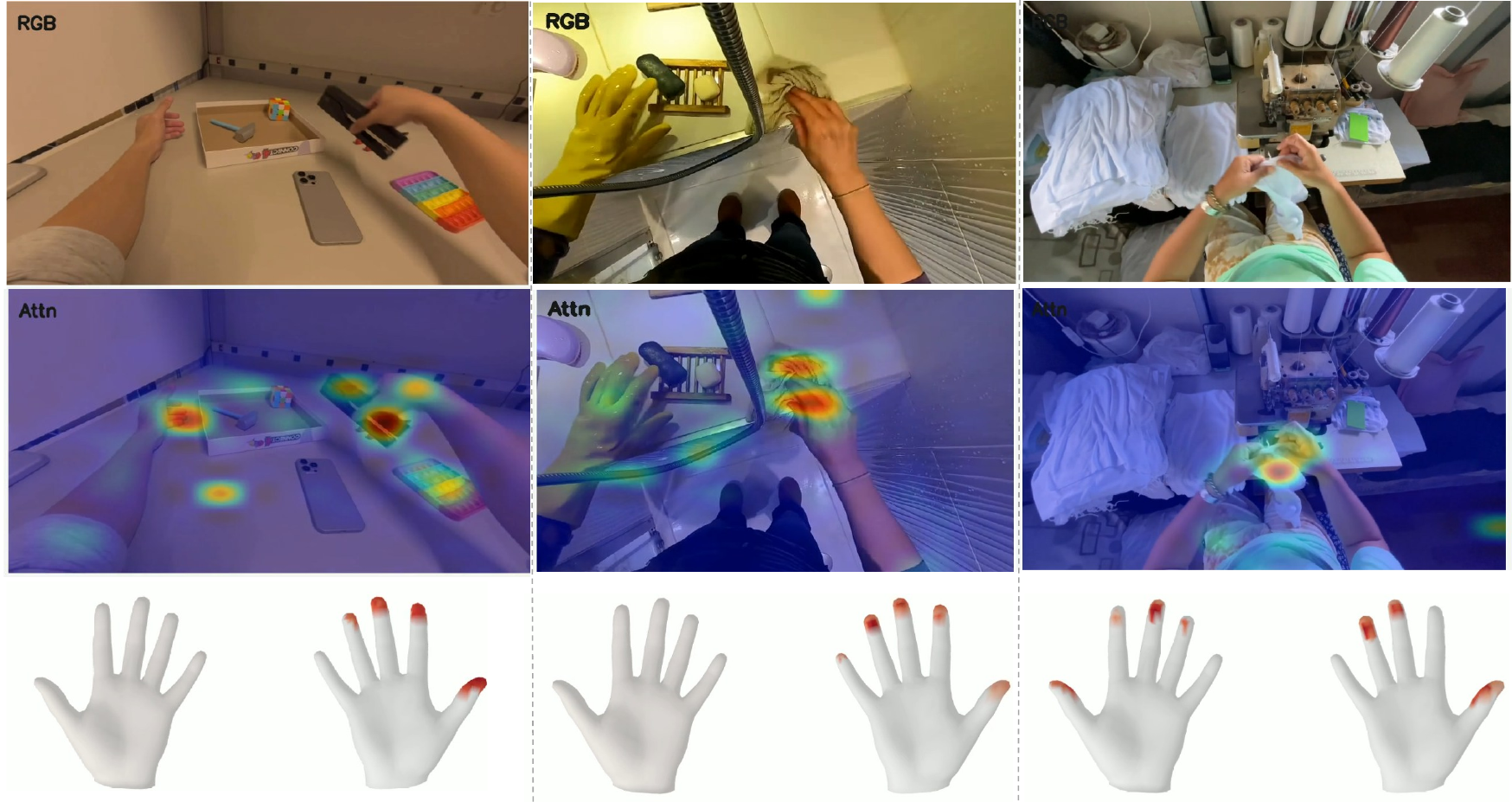}
\end{center}
\vspace{-0.1cm}
\caption{
\textbf{Attention maps from the DINO encoder~\cite{oquab2023dinov2}.} Across diverse tasks, EgoTac emphasizes hand-object interaction regions that are informative for tactile prediction.}
\label{fig:supp_attention}
\vspace{-0.1cm}
\end{figure*}

\begin{figure*}[t]
\begin{center}
\includegraphics[width=1\linewidth]{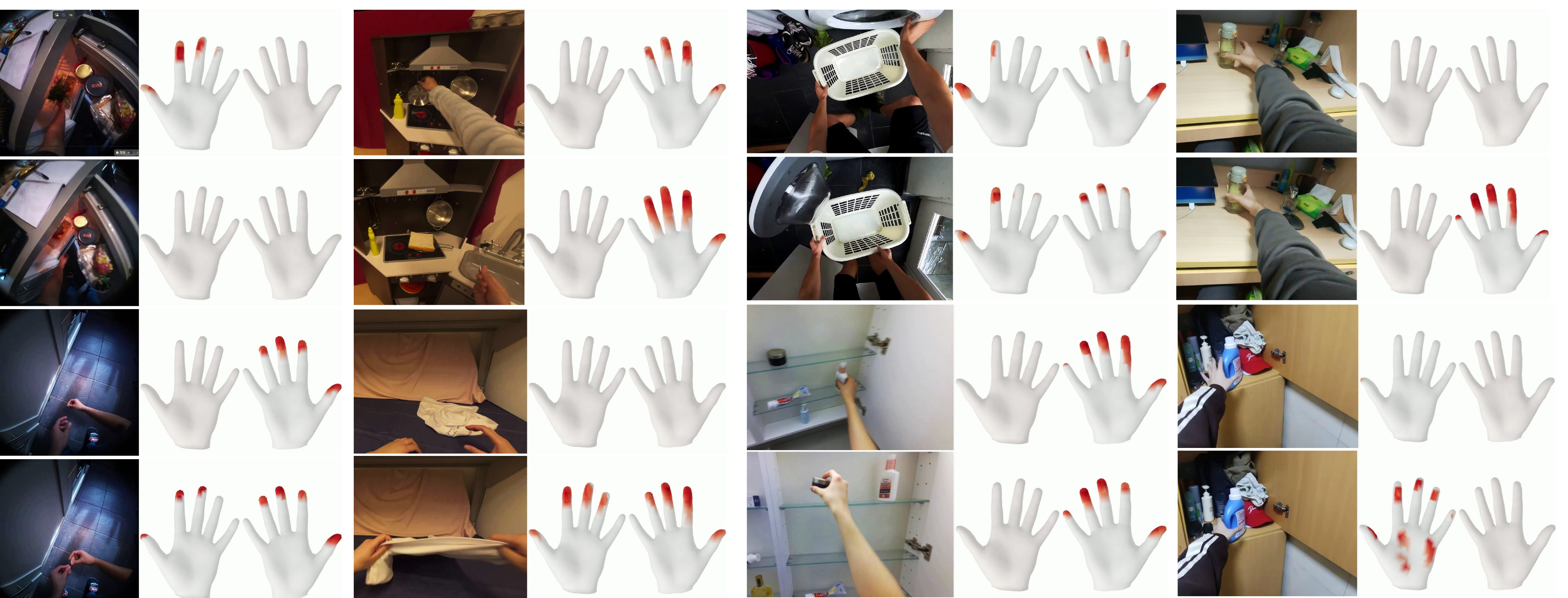}
\end{center}
\vspace{-0.1cm}
\caption{
\textbf{Additional qualitative results in diverse in-the-wild scenes.} We show predictions on EPIC-KITCHENS~\cite{damen2018scaling}, EgoDex~\cite{hoque2025egodex}, Ego4D~\cite{grauman2022ego4d}, EgoPAT3D~\cite{li2022egocentric}, and self-collected videos.}
\label{fig:supp_wild}
\vspace{-0.1cm}
\end{figure*}

\section{Details of metrics}
\label{sec:supp_metrics}

This section defines the metrics reported in Sec.~\ref{sec:experiments}. We group them into four categories: (1) in-domain continuous force metrics, (2) contact benchmark metrics, (3) OpenTouch OOD force-dynamics metrics, and (4) frequency-domain force-dynamics metrics.

\subsection{Tactile-related metrics}
\label{sec:supp_metrics_tactile}

For in-domain continuous tactile evaluation, the main paper reports three metrics: global MAE, active-region MAE and temporal Pearson correlation.

Let \(F_{t,v}^{\mathrm{pred}}\) and \(F_{t,v}^{\mathrm{gt}}\) denote predicted and ground-truth force at frame \(t\) and MANO vertex \(v\), and let \(\Omega\) be the valid evaluation set.

\paragraph{Global MAE.}
\begin{equation}
\mathrm{MAE}=\frac{1}{|\Omega|}\sum_{(t,v)\in\Omega}\left|F_{t,v}^{\mathrm{pred}}-F_{t,v}^{\mathrm{gt}}\right|,
\end{equation}
which measures overall force error across the full hand surface. This metric is sensitive to the strong class imbalance between contact and non-contact regions.

\paragraph{Active-region MAE.}
We define active vertices by a force threshold \(\theta=0.05\) N:
\begin{equation}
\Omega_{\mathrm{act}}=\{(t,v)\in\Omega \mid F_{t,v}^{\mathrm{gt}}>\theta\}.
\end{equation}
Then
\begin{equation}
\mathrm{MAE}_{\mathrm{act}}=\frac{1}{|\Omega_{\mathrm{act}}|}\sum_{(t,v)\in\Omega_{\mathrm{act}}}\left|F_{t,v}^{\mathrm{pred}}-F_{t,v}^{\mathrm{gt}}\right|,
\end{equation}
which focuses on physically meaningful contact areas and is therefore emphasized in our analysis.


\paragraph{Temporal Pearson correlation.}
For each frame, we compute total hand force
\begin{equation}
S_t = \sum_{v} F_{t,v}.
\end{equation}
Temporal Pearson is
\begin{equation}
\mathrm{TempPearson}=\rho\!\left(\{S_t^{\mathrm{pred}}\}_t,\{S_t^{\mathrm{gt}}\}_t\right),
\end{equation}
which measures whether the predicted force trajectory follows the temporal evolution of real interaction intensity.

\subsection{Contact-related metrics}
\label{sec:supp_metrics_contact}

For contact benchmarks (in-domain and OOD), the main paper reports Precision, Recall, F1, IoU, and AUROC for MANO contact prediction.

Given binary prediction \(\hat{y}\) and ground truth \(y\), with TP/FP/FN/TN defined at vertex level:
\begin{equation}
\mathrm{Precision}=\frac{\mathrm{TP}}{\mathrm{TP}+\mathrm{FP}},\quad
\mathrm{Recall}=\frac{\mathrm{TP}}{\mathrm{TP}+\mathrm{FN}},
\end{equation}
\begin{equation}
\mathrm{F1}=\frac{2\cdot \mathrm{Precision}\cdot \mathrm{Recall}}{\mathrm{Precision}+\mathrm{Recall}},\quad
\mathrm{IoU}=\frac{\mathrm{TP}}{\mathrm{TP}+\mathrm{FP}+\mathrm{FN}},
\end{equation}
and AUROC is computed from continuous contact scores before thresholding.
Precision emphasizes false-positive control, Recall emphasizes missed-contact control, F1 balances the two, and IoU provides a stricter overlap measure under sparse positives.

\subsection{OpenTouch OOD force-dynamics metrics}
\label{sec:supp_metrics_opentouch}

For the OOD force-sensitive benchmark on OpenTouch~\cite{song2025opentouch}, we report \textbf{force-cosine}, \textbf{rise F1}, and \textbf{fall F1}, as in Sec.~\ref{sec:ood_eval}. Since sensor calibration differs across datasets, we evaluate relative dynamics after normalizing force signals to \([0,1]\).

\paragraph{Force-cosine.}
Let \(\mathbf{s}^{\mathrm{pred}},\mathbf{s}^{\mathrm{gt}}\in\mathbb{R}^{T}\) be predicted and ground-truth total-force time series (after normalization and masking). Force-cosine is
\begin{equation}
\mathrm{Force\mbox{-}Cosine}
=
\frac{\langle \mathbf{s}^{\mathrm{pred}},\mathbf{s}^{\mathrm{gt}} \rangle}
{\|\mathbf{s}^{\mathrm{pred}}\|_2\|\mathbf{s}^{\mathrm{gt}}\|_2+\epsilon}.
\end{equation}
It measures trend-shape consistency independent of global scale.

\paragraph{Rise F1 and Fall F1.}
Define first-order differences
\begin{equation}
\Delta s_t = s_t - s_{t-1}.
\end{equation}
With event threshold \(\eta = 1e^{-4}>0\), rise and fall events are:
\begin{equation}
e_t^{\uparrow}=\mathbbm{1}[\Delta s_t>\eta],\qquad
e_t^{\downarrow}=\mathbbm{1}[\Delta s_t<-\eta].
\end{equation}
We compute F1 between predicted and ground-truth event sequences separately:
\begin{equation}
\mathrm{Rise\ F1}= \mathrm{F1}(e^{\uparrow,\mathrm{pred}},e^{\uparrow,\mathrm{gt}}),\quad
\mathrm{Fall\ F1}= \mathrm{F1}(e^{\downarrow,\mathrm{pred}},e^{\downarrow,\mathrm{gt}}).
\end{equation}

These two metrics evaluate whether the model captures \emph{when force increases} and \emph{when force decreases}, which is crucial for contact transition dynamics and manipulation timing.

\subsection{Frequency-domain metrics}
\label{sec:supp_frequency_metrics}

In addition to the time-domain metrics above, we evaluate whether the
predicted tactile signals preserve the temporal structure of real
interactions in the frequency domain. These metrics provide a
complementary characterization of tactile dynamics beyond point-wise
magnitude errors and local rise/fall events.

For each episode, we first aggregate the dense MANO tactile field into
a scalar force trace:
\begin{equation}
    s_t = \sum_v F_{t,v}.
\end{equation}
Given a force sequence $\mathbf{s}=\{s_t\}_{t=1}^{T}$, we remove its
DC component and apply a Hann window before computing a single-sided
real discrete Fourier transform (DFT):
\begin{equation}
    \widetilde{s}_t
    =
    w_t
    \left(
        s_t - \frac{1}{T}\sum_{\tau=1}^{T}s_\tau
    \right),
    \qquad
    \mathcal{F}
    =
    \operatorname{rFFT}(\widetilde{\mathbf{s}}),
\end{equation}
where $w_t$ is the Hann-window coefficient. We apply the same procedure
to the predicted and ground-truth force traces and denote their spectra
as $\mathcal{F}^{\mathrm{pred}}$ and $\mathcal{F}^{\mathrm{gt}}$,
respectively.

\paragraph{DFT log-magnitude Pearson correlation.}
We first compute the log-magnitude spectrum
\begin{equation}
    A(f)
    =
    \log\left(1+\left|\mathcal{F}(f)\right|\right).
\end{equation}
The DFT-Pearson metric is then defined as
\begin{equation}
    \mathrm{DFT\text{-}Pearson}
    =
    \rho\left(
        A^{\mathrm{pred}},
        A^{\mathrm{gt}}
    \right),
\end{equation}
where $\rho(\cdot,\cdot)$ denotes the Pearson correlation across
frequency bins. This metric measures the similarity between the
predicted and ground-truth spectral envelopes while being relatively
insensitive to a global amplitude scale. Higher is better.

\paragraph{Spectral convergence.}
To additionally measure spectral magnitude mismatch, we compute
\begin{equation}
    \mathrm{SpecConv}
    =
    \frac{
        \left\|
        |\mathcal{F}^{\mathrm{pred}}|
        -
        |\mathcal{F}^{\mathrm{gt}}|
        \right\|_2
    }{
        \left\|
        |\mathcal{F}^{\mathrm{gt}}|
        \right\|_2 + \epsilon
    }.
\end{equation}
Unlike DFT-Pearson, spectral convergence is sensitive to differences in
the magnitude of frequency components. Lower is better.

\paragraph{Dominant-frequency error.}
We further compare the locations of the strongest non-DC frequency
components. Let
\begin{equation}
    f_{\mathrm{dom}}
    =
    \arg\max_{f>0}
    \left|\mathcal{F}(f)\right|.
\end{equation}
The dominant-frequency error is defined as
\begin{equation}
    E_{\mathrm{dom}}
    =
    \left|
        f_{\mathrm{dom}}^{\mathrm{pred}}
        -
        f_{\mathrm{dom}}^{\mathrm{gt}}
    \right|.
\end{equation}
This metric measures whether the dominant interaction rhythm of the
predicted tactile signal occurs at a frequency similar to that of the
ground truth. We report the error in Hz, and lower is better.

Together, these frequency-domain metrics evaluate whether EgoTac
preserves the spectral structure of tactile dynamics, complementing
the time-domain force-cosine and rise/fall metrics.

\section{More experimental results}
\label{sec:supp_more_results}

This section provides additional results omitted from the main paper due to space constraints.

\paragraph{Frequency-domain evaluation.}
To complement the time-domain force-dynamics metrics, we further
evaluate the spectral fidelity of predicted tactile signals using the
frequency-domain metrics defined in
Sec.~\ref{sec:supp_frequency_metrics}, including DFT log-magnitude
Pearson correlation, spectral convergence, and dominant-frequency error.

We first evaluate on the in-domain EgoTac-SC test set.
As shown in Table~\ref{tab:supp_indomain_frequency}, EgoTac consistently
outperforms the image-based EgoTac-i variant. In particular,
DFT-Pearson improves from 0.831 to 0.860, while dominant-frequency
error decreases from 0.206 Hz to 0.184 Hz, indicating that temporal
visual context improves both time-domain and spectral tactile dynamics.

We further evaluate OOD transfer on OpenTouch under different
training-data scales. Table~\ref{tab:supp_opentouch_frequency} shows
that EgoTac consistently preserves the coarse spectral structure of
the ground-truth tactile trajectories. DFT-Pearson ranges from
0.744 to 0.771 and dominant-frequency error remains between
0.108 and 0.131 Hz. With the full training set, EgoTac achieves the
best DFT-Pearson of 0.771 and spectral convergence of 0.641, together
with the strongest overall temporal-dynamics performance.

Together, these results show that EgoTac captures tactile dynamics in
both the temporal and frequency domains. On EgoTac-SC, calibrated MAE
measures force-magnitude accuracy within the same sensing system,
whereas the temporal and spectral metrics characterize dynamics
fidelity. On OpenTouch, where the sensing hardware and calibration
differ, these metrics evaluate cross-sensor transfer of normalized
tactile dynamics rather than Newton-calibrated absolute-force accuracy.

\begin{table}[t]
    \centering
    \small
    \setlength{\tabcolsep}{5pt}
    \caption{
    Temporal and frequency-domain evaluation on the in-domain
    EgoTac-SC evaluation set.
    }
    \label{tab:supp_indomain_frequency}
    \begin{tabular}{lcccccc}
        \toprule
        Method
        & \shortstack{Force-\\Cosine $\uparrow$}
        & \shortstack{Rise-\\F1 $\uparrow$}
        & \shortstack{Fall-\\F1 $\uparrow$}
        & \shortstack{DFT-\\Pearson $\uparrow$}
        & \shortstack{Spec.\\Conv. $\downarrow$}
        & \shortstack{Domin. Freq.\\Error $\downarrow$} \\
        \midrule
        EgoTac-i
        & 0.829 & 0.396 & 0.391 & 0.831 & 0.684 & 0.206 \\
        EgoTac
        & \textbf{0.844} & \textbf{0.424} & \textbf{0.397}
        & \textbf{0.860} & \textbf{0.645} & \textbf{0.184} \\
        \bottomrule
    \end{tabular}
\end{table}
\begin{table}[t]
    \centering
    \small
    \setlength{\tabcolsep}{5pt}
    \caption{
    Temporal and frequency-domain evaluation on the OOD OpenTouch
    benchmark under different training-data scales. (We use EgoTac-i for this ablation)
    }
    \label{tab:supp_opentouch_frequency}
    \begin{tabular}{ccccccc}
        \toprule
        \shortstack{Data Use\\Ratio}
        & \shortstack{Force-\\Cosine $\uparrow$}
        & \shortstack{Rise-\\F1 $\uparrow$}
        & \shortstack{Fall-\\F1 $\uparrow$}
        & \shortstack{DFT-\\Pearson $\uparrow$}
        & \shortstack{Spec.\\Conv. $\downarrow$}
        & \shortstack{Domin. Freq.\\Error $\downarrow$} \\
        \midrule
        10\%
        & 0.754 & 0.397 & 0.394 & 0.754 & 0.683 & 0.127 \\
        20\%
        & 0.756 & 0.408 & 0.409 & 0.767 & 0.656 & \textbf{0.108} \\
        50\%
        & 0.756 & 0.413 & 0.405 & 0.744 & 0.699 & 0.131 \\
        100\%
        & \textbf{0.787} & \textbf{0.419} & \textbf{0.417}
        & \textbf{0.771} & \textbf{0.641} & 0.119 \\
        \bottomrule
    \end{tabular}
\end{table}

\paragraph{EgoTac-SC dataset ablation.}
We conduct additional ablations to isolate the contribution of the
self-collected EgoTac-SC dataset from the existing contact-only HOI
datasets. Specifically, we compare three training configurations:
(1) EgoTac-SC only, (2) contact-only HOI datasets only, and
(3) the full mixed-data setting. Here, the contact-only HOI datasets
include HOT3D, HOI4D, TACO, H2O, and ARCTIC. The mixed setting,
EgoTac-SC + Contact-only HOI, is exactly the training configuration
used by the full EgoTac model and does not introduce any additional
training data.

We first evaluate OOD contact prediction on OAKINK2 using the same
whole-hand protocol as the main paper, where all 778 MANO vertices are
included in evaluation. As shown in Table~\ref{tab:supp_dataset_ablation_oakink},
training only on EgoTac-SC results in substantially weaker OOD contact
generalization, whereas the contact-only HOI datasets provide strong
cross-domain contact performance. The mixed model maintains this strong
contact generalization while additionally benefiting from the continuous
force supervision provided by EgoTac-SC.

We further evaluate the same training configurations on the OOD
force-sensitive OpenTouch benchmark. As shown in
Table~\ref{tab:supp_dataset_ablation_opentouch}, contact-only HOI data
cannot independently train a continuous force predictor because it
contains only binary contact supervision. In contrast, EgoTac-SC
provides the continuous tactile annotations required for learning
force dynamics. Combining EgoTac-SC with contact-only HOI data retains
comparable force-dynamics performance while improving several metrics,
suggesting that the two supervision sources are complementary.

\begin{table}[t]
    \centering
    \small
    \setlength{\tabcolsep}{8pt}
    \caption{
    Dataset ablation on the OOD contact benchmark OAKINK2.
    All results use the same whole-hand evaluation protocol as the
    cross-method comparison in the main paper.
    }
    \label{tab:supp_dataset_ablation_oakink}
    \begin{tabular}{lccccc}
        \toprule
        Training setting
        & AUROC $\uparrow$
        & IoU $\uparrow$
        & Precision $\uparrow$
        & Recall $\uparrow$
        & F1 $\uparrow$ \\
        \midrule
        EgoTac-SC only
        & 0.60 & 0.07 & 0.35 & 0.08 & 0.12 \\
        Contact-only HOI
        & \textbf{0.84} & \textbf{0.29} & \textbf{0.51}
        & \textbf{0.45} & \textbf{0.44} \\
        EgoTac-SC + Contact-only HOI
        & \textbf{0.84} & \textbf{0.29} & 0.50
        & 0.44 & \textbf{0.44} \\
        \bottomrule
    \end{tabular}
\end{table}

\begin{table}[t]
    \centering
    \small
    \setlength{\tabcolsep}{4pt}
    \caption{
    Dataset ablation on the OOD force-dynamics benchmark OpenTouch.
    Contact-only HOI datasets do not contain continuous force targets
    and therefore cannot independently train the force predictor.
    }
    \label{tab:supp_dataset_ablation_opentouch}
    \begin{tabular}{lcccccc}
        \toprule
        Training setting
        & \shortstack{Force-\\Cosine $\uparrow$}
        & \shortstack{Rise-\\F1 $\uparrow$}
        & \shortstack{Fall-\\F1 $\uparrow$}
        & \shortstack{DFT-\\Pearson $\uparrow$}
        & \shortstack{Spec.\\Conv. $\downarrow$}
        & \shortstack{Domin. Freq.\\Error $\downarrow$} \\
        \midrule
        EgoTac-SC only
        & 0.745 & \textbf{0.404} & 0.385
        & \textbf{0.767} & \textbf{0.656} & 0.142 \\
        Contact-only HOI
        & -- & -- & -- & -- & -- & -- \\
        EgoTac-SC + Contact-only HOI
        & \textbf{0.777} & 0.403 & \textbf{0.403}
        & 0.756 & 0.693 & \textbf{0.127} \\
        \bottomrule
    \end{tabular}
\end{table}

\begin{table}[t]
    \centering
    \small
    \setlength{\tabcolsep}{4pt}
    \caption{
    Ablation of the force--contact consistency loss
    $\mathcal{L}_{\mathrm{cons}}$ on the OOD OpenTouch benchmark.
    }
    \label{tab:supp_lcons_ablation}
    \begin{tabular}{lcccccc}
        \toprule
        Training setting
        & \shortstack{Force-\\Cosine $\uparrow$}
        & \shortstack{Rise-\\F1 $\uparrow$}
        & \shortstack{Fall-\\F1 $\uparrow$}
        & \shortstack{DFT-\\Pearson $\uparrow$}
        & \shortstack{Spec.\\Conv. $\downarrow$}
        & \shortstack{Domin. Freq.\\Error $\downarrow$} \\
        \midrule

        EgoTac w/o $\mathcal{L}_{\mathrm{cons}}$
        & 0.759
        & 0.398
        & 0.400
        & 0.753
        & \textbf{0.672}
        & 0.131 \\

        EgoTac
        & \textbf{0.777}
        & \textbf{0.403}
        & \textbf{0.403}
        & \textbf{0.756}
        & 0.693
        & \textbf{0.127} \\
        \bottomrule
    \end{tabular}
\end{table}

\begin{table}[t]
\centering
\small
\caption{\textbf{More results of in-domain contact evaluation.} Results are averaged across timestamps on all test splits of each component dataset.}
\label{tab:supp_id_contact}
\begin{tabular}{llccccc}
\toprule
Benchmark & Method & AUROC $\uparrow$ & IoU $\uparrow$ & Precision $\uparrow$ & Recall $\uparrow$ & F1 $\uparrow$ \\
\midrule
\multirow{2}{*}{TACO~\cite{liu2024taco}} 
& EgoTac-i & \textbf{0.974} & \textbf{0.657} & \textbf{0.732} & 0.855 & \textbf{0.788} \\
& \textbf{EgoTac} & 0.973 & 0.656 & 0.729 & \textbf{0.858} & 0.787 \\
\midrule
\multirow{2}{*}{HOI4D~\cite{liu2022hoi4d}} 
& EgoTac-i & \textbf{0.961} & 0.575 & 0.647 & \textbf{0.828} & 0.720 \\
& \textbf{EgoTac} & 0.959 & \textbf{0.579} & \textbf{0.657} & 0.818 & \textbf{0.724} \\
\bottomrule
\end{tabular}
\vspace{-0.4cm}
\end{table}
\begin{table}[t]
\centering
\small
\caption{\textbf{Data volume scaling on in-domain contact estimation.} As training data scales from 10\% to 100 \%, contact estimation performance on held-out in-domain test sets consistently improves. We conduct these experiments on EgoTac-i.}
\label{tab:supp_data_scale}
\begin{tabular}{lcccccc}
\toprule
Benchmark & Use Ratio & AUROC $\uparrow$ & IoU $\uparrow$ & Precision $\uparrow$ & Recall $\uparrow$ & F1 $\uparrow$ \\
\midrule

\multirow{4}{*}{HOT3D~\cite{banerjee2025hot3d}} 
& 10\% & 0.914 & 0.392 & 0.586 & 0.524 & 0.524 \\
& 20\% & 0.932 & 0.454 & 0.603 & 0.603 & 0.581 \\
& 50\% & 0.958 & 0.539 & 0.634 & \textbf{0.771} & 0.658 \\
& 100\% & \textbf{0.976} & \textbf{0.587 }& \textbf{0.656} & \textbf{0.771} & \textbf{0.702} \\

\midrule

\multirow{4}{*}{H2O~\cite{kwon2021h2o}} 
& 10\% & 0.936 & 0.323 & 0.565 & 0.469 & 0.458 \\
& 20\% & 0.954 & 0.425 & 0.626 & 0.613 & 0.574 \\
& 50\% & 0.975 & 0.567 & 0.708 & 0.737 & 0.706 \\
& 100\% & \textbf{0.989} & \textbf{0.647} & \textbf{0.725} & \textbf{0.846} & \textbf{0.777} \\

\midrule

\multirow{4}{*}{ARCTIC~\cite{fan2023arctic}} 
& 10\% & 0.810 & 0.236 & 0.462 & 0.356 & 0.381 \\
& 20\% & 0.846 & 0.330 & 0.483 & 0.518 & 0.480 \\
& 50\% & 0.910 & 0.475 & 0.628 & 0.674 & 0.636 \\
& 100\% & \textbf{0.972} & \textbf{0.676} & \textbf{0.716} & \textbf{0.905} & \textbf{0.790} \\

\midrule

\multirow{4}{*}{TACO~\cite{liu2024taco}} 
& 10\% & 0.933 & 0.510 & 0.647 & 0.689 & 0.660 \\
& 20\% & 0.949 & 0.564 & 0.686 & 0.743 & 0.708 \\
& 50\% & 0.968 & 0.634 & 0.720 & 0.830 & 0.769 \\
& 100\% & \textbf{0.974} &\textbf{ 0.657} & \textbf{0.732} & \textbf{0.855} & \textbf{0.788} \\

\midrule

\multirow{4}{*}{HOI4D~\cite{liu2022hoi4d}} 
& 10\% & 0.934 & 0.449 & 0.638 & 0.629 & 0.600 \\
& 20\% & 0.943 & 0.493 & 0.641 & 0.700 & 0.644 \\
& 50\% & 0.955 & 0.547 & 0.644 & 0.785 & 0.693 \\
& 100\% & \textbf{0.961} & \textbf{0.575} & \textbf{0.647} & \textbf{0.828} & \textbf{0.720} \\

\bottomrule
\end{tabular}
\vspace{-0.4cm}
\end{table}

Overall, these results highlight the complementary roles of the two
data sources. EgoTac-SC uniquely provides \textbf{real continuous-force
supervision} that enables tactile-intensity and force-dynamics learning,
while heterogeneous bare-hand contact datasets substantially improve
\textbf{visual diversity and OOD contact generalization}.

\paragraph{$\mathcal{L}_{\mathrm{cons}}$ ablation.}
To further understand why contact-only supervision can affect force
prediction, we isolate the contribution of the force--contact consistency
loss $\mathcal{L}_{\mathrm{cons}}$. Specifically, we train an additional
mixed-data model using the same EgoTac-SC and contact-only HOI datasets,
while removing $\mathcal{L}_{\mathrm{cons}}$ and keeping all other
training settings unchanged.

As shown in Table~\ref{tab:supp_lcons_ablation}, mixed-source training
without $\mathcal{L}_{\mathrm{cons}}$ already improves several
force-dynamics metrics over EgoTac-SC-only training. For example,
force-cosine increases from 0.745 to 0.759 and dominant-frequency error
decreases from 0.142 to 0.131 Hz. This suggests that the broader visual
coverage provided by contact-only HOI data can benefit force prediction
through the shared visual-temporal representation, even without an
explicit consistency constraint.

Adding $\mathcal{L}_{\mathrm{cons}}$ further improves force-cosine from
0.759 to 0.777, fall F1 from 0.400 to 0.403, and dominant-frequency
error from 0.131 to 0.127 Hz. However, the improvement is not consistent
across all metrics. We therefore interpret $\mathcal{L}_{\mathrm{cons}}$
as providing a modest structural regularization effect rather than being
the primary source of the mixed-data improvement.

\paragraph{Additional in-domain results.} We report additional in-domain evaluation results on TACO~\cite{liu2024taco} and HOI4D~\cite{liu2022hoi4d} test sets in Table~\ref{tab:supp_id_contact}, along with additional results for training-data volume scaling on in-domain contact benchmarks in Table~\ref{tab:supp_data_scale}.

\paragraph{Additional visualizations.} Figure~\ref{fig:supp_attention} presents attention maps in diverse scenes, showing that EgoTac consistently focuses on hand-object interaction regions that determine tactile patterns.
Figure~\ref{fig:supp_wild} provides additional qualitative examples of in-the-wild tactile prediction from egocentric videos.

\section{More discussions}
\label{sec:supp_more_discussions}

This section provides further discussion on several aspects of EgoTac,
including the visual-domain gap introduced by tactile gloves, the
interaction between contact-only supervision and force prediction, and
the scope of our architectural contribution.

\paragraph{Visual-domain gap introduced by tactile gloves.}
EgoTac-SC is collected using wearable tactile gloves, which inevitably
introduces a visual appearance gap between the force-supervised training
data and unconstrained bare-hand videos. A natural strategy is to remove
or translate the glove appearance before training. In a preliminary
study, we used Grounded SAM-2 to segment the gloves and recolored the
segmented regions toward a skin-like appearance. However, we did not
observe a clear improvement in zero-shot tactile prediction on
bare-hand videos.

We believe that simple recoloring or naive inpainting is insufficient
for this problem. The glove physically occludes the underlying hand
appearance, and therefore appearance translation must reconstruct
rather than merely recolor the missing visual information. Moreover,
artifacts around finger boundaries, hand--object occlusions, and
fine-grained contact regions may remove or distort precisely the visual
cues that are most informative for tactile prediction.

Instead, EgoTac mitigates this domain gap through heterogeneous
mixed-source training. We map multiple bare-hand HOI datasets into the
same MANO-aligned representation as EgoTac-SC and jointly train the
model using gloved force-supervised data and bare-hand contact-supervised
data. As shown by the dataset ablations in
Sec.~\ref{sec:supp_more_results}, the bare-hand contact datasets
substantially improve OOD contact generalization, while EgoTac-SC
provides continuous-force supervision that is unavailable in these
contact-only datasets. We therefore view mixed-domain supervision as a
complementary alternative to direct appearance translation.
More sophisticated geometry-preserving glove-to-hand translation~\cite{zhang2026glove2hand}
remains a promising direction for future work.

\paragraph{Why contact-only data can affect force dynamics.}
Contact-only HOI samples contain no continuous-force targets.
Accordingly, their force-valid mask $M_F$ is zero, and they do not
directly contribute to the force-regression loss $\mathcal{L}_F$ or
the active-region loss $\mathcal{L}_A$. Nevertheless, contact-only
data can still influence force prediction through two complementary
mechanisms.

First, the force and contact tasks share the same visual encoder,
temporal fusion module, and tactile decoder. The contact classification
objective therefore improves the shared visual-temporal representation
using a broader distribution of bare-hand appearances, objects,
viewpoints, scenes, and contact transitions. These improved shared
features can subsequently benefit the force prediction branch even
without direct force supervision.

Second, the consistency objective $\mathcal{L}_{\mathrm{cons}}$
provides an explicit interaction between the two prediction branches.
As described in Sec.~\ref{sec:supp_impl_loss},
$\mathcal{L}_{\mathrm{cons}}$ aligns the contact logits implicitly
derived from the predicted force with the detached prediction of the
contact head. Therefore, for contact-only samples, the contact
prediction can provide the force branch with a binary structural
constraint indicating \emph{where} and \emph{when} force should be
active, although it does not specify the corresponding force magnitude.

The controlled ablation in Sec.~\ref{sec:supp_more_results} further
shows that removing $\mathcal{L}_{\mathrm{cons}}$ from mixed-data
training only moderately changes the OpenTouch force-dynamics metrics.
This suggests that the benefit of contact-only data cannot be attributed
to the consistency loss alone. Instead, we interpret the improvement as
a combination of broader visual-domain coverage through shared
representation learning and a smaller regularization effect from
$\mathcal{L}_{\mathrm{cons}}$. Importantly, continuous tactile
intensity itself remains uniquely supervised by EgoTac-SC.

\paragraph{Architectural novelty.}
The individual architectural components of EgoTac, including the
pretrained vision backbone, temporal attention, and AdaLN-based
Transformer decoder, are established techniques. Our contribution is
therefore not intended to introduce a new Transformer primitive or a
highly specialized network module. Instead, the technical contribution
lies in formulating and enabling dense vision-to-tactile learning from
large-scale heterogeneous supervision.

Specifically, EgoTac represents continuous tactile force and binary
contact on a shared two-hand MANO topology, allowing annotations from
different sources to share a common output space. It jointly learns
from force-supervised tactile data and contact-only HOI datasets despite
their different annotation availability, using force-valid and
contact-valid masks so that each sample supervises only the annotations
it provides. The force and contact predictions are further coupled by
the consistency objective, which encourages structural agreement between
continuous tactile intensity and binary contact.

Together, these design choices allow data sources with fundamentally
different label types, spatial coverage, and visual domains to contribute
to a single dense tactile predictor. We intentionally adopt a concise
encoder--decoder architecture so that the effects of the unified
representation, heterogeneous supervision, and data scaling can be
studied without conflating them with gains from a highly specialized
backbone. We therefore view the main technical novelty of EgoTac as the
unified tactile representation and heterogeneous force--contact learning
framework, rather than the introduction of a new backbone architecture.

\section{Limitations and societal impacts}
\label{sec:supp_limitations}

\paragraph{Limitations.} Despite these advances, EgoTac has limitations: it may underperform under severe occlusion, strong motion blur or rare interaction patterns. Future work could address these issues by incorporating occlusion reasoning and motion deblurring, as well as expanding training data diversity to better capture rare interactions. In addition, EgoTac-SC is collected with tactile gloves, introducing
a visual-domain gap with bare-hand videos. Beyond methodological improvements, EgoTac-predicted tactile signals offer rich priors for downstream applications, such as pretraining robotic manipulation policies, enhancing simulation realism, and enabling contact-aware action planning in human-robot interaction. These directions collectively advance the integration of visual perception and tactile understanding for more physically informed AI and robotic systems.

\paragraph{Societal impacts.}This work can benefit robotics, AR/VR, and assistive systems by enabling richer hand-state estimation from egocentric visual input. Potential misuse includes privacy-invasive behavior inference from first-person recordings. Responsible deployment should require explicit consent, strict data governance, and task-specific risk assessment before real-world use.

\clearpage
\section*{NeurIPS Paper Checklist}

\begin{enumerate}

\item {\bf Claims}
    \item[] Question: Do the main claims made in the abstract and introduction accurately reflect the paper's contributions and scope?
    \item[] Answer: \answerYes{} 
    \item[] Justification: The claims in the abstract and introduction correspond directly to the methodology (Section 4) and are empirically supported by the experiments (Section 5).
    \item[] Guidelines:
    \begin{itemize}
        \item The answer \answerNA{} means that the abstract and introduction do not include the claims made in the paper.
        \item The abstract and/or introduction should clearly state the claims made, including the contributions made in the paper and important assumptions and limitations. A \answerNo{} or \answerNA{} answer to this question will not be perceived well by the reviewers. 
        \item The claims made should match theoretical and experimental results, and reflect how much the results can be expected to generalize to other settings. 
        \item It is fine to include aspirational goals as motivation as long as it is clear that these goals are not attained by the paper. 
    \end{itemize}

\item {\bf Limitations}
    \item[] Question: Does the paper discuss the limitations of the work performed by the authors?
    \item[] Answer: \answerYes{} 
    \item[] Justification: Appendices~\ref{sec:supp_more_discussions} and~\ref{sec:supp_limitations} provide discussions on the visual-domain gap, difficult visual conditions, and the scope of force and downstream evaluations.
    \item[] Guidelines:
    \begin{itemize}
        \item The answer \answerNA{} means that the paper has no limitation while the answer \answerNo{} means that the paper has limitations, but those are not discussed in the paper. 
        \item The authors are encouraged to create a separate ``Limitations'' section in their paper.
        \item The paper should point out any strong assumptions and how robust the results are to violations of these assumptions (e.g., independence assumptions, noiseless settings, model well-specification, asymptotic approximations only holding locally). The authors should reflect on how these assumptions might be violated in practice and what the implications would be.
        \item The authors should reflect on the scope of the claims made, e.g., if the approach was only tested on a few datasets or with a few runs. In general, empirical results often depend on implicit assumptions, which should be articulated.
        \item The authors should reflect on the factors that influence the performance of the approach. For example, a facial recognition algorithm may perform poorly when image resolution is low or images are taken in low lighting. Or a speech-to-text system might not be used reliably to provide closed captions for online lectures because it fails to handle technical jargon.
        \item The authors should discuss the computational efficiency of the proposed algorithms and how they scale with dataset size.
        \item If applicable, the authors should discuss possible limitations of their approach to address problems of privacy and fairness.
        \item While the authors might fear that complete honesty about limitations might be used by reviewers as grounds for rejection, a worse outcome might be that reviewers discover limitations that aren't acknowledged in the paper. The authors should use their best judgment and recognize that individual actions in favor of transparency play an important role in developing norms that preserve the integrity of the community. Reviewers will be specifically instructed to not penalize honesty concerning limitations.
    \end{itemize}

\item {\bf Theory assumptions and proofs}
    \item[] Question: For each theoretical result, does the paper provide the full set of assumptions and a complete (and correct) proof?
    \item[] Answer: \answerNA{} 
    \item[] Justification: This paper does not present theoretical proofs, as it focuses on empirical methodology and dataset construction.
    \item[] Guidelines:
    \begin{itemize}
        \item The answer \answerNA{} means that the paper does not include theoretical results. 
        \item All the theorems, formulas, and proofs in the paper should be numbered and cross-referenced.
        \item All assumptions should be clearly stated or referenced in the statement of any theorems.
        \item The proofs can either appear in the main paper or the supplemental material, but if they appear in the supplemental material, the authors are encouraged to provide a short proof sketch to provide intuition. 
        \item Inversely, any informal proof provided in the core of the paper should be complemented by formal proofs provided in appendix or supplemental material.
        \item Theorems and Lemmas that the proof relies upon should be properly referenced. 
    \end{itemize}

    \item {\bf Experimental result reproducibility}
    \item[] Question: Does the paper fully disclose all the information needed to reproduce the main experimental results of the paper to the extent that it affects the main claims and/or conclusions of the paper (regardless of whether the code and data are provided or not)?
    \item[] Answer: \answerYes{} 
    \item[] Justification: We describe our model architecture explicitly in Section 4 and detail the complete training protocols and evaluation metrics in the supplementary material.
    \item[] Guidelines:
    \begin{itemize}
        \item The answer \answerNA{} means that the paper does not include experiments.
        \item If the paper includes experiments, a \answerNo{} answer to this question will not be perceived well by the reviewers: Making the paper reproducible is important, regardless of whether the code and data are provided or not.
        \item If the contribution is a dataset and\slash or model, the authors should describe the steps taken to make their results reproducible or verifiable. 
        \item Depending on the contribution, reproducibility can be accomplished in various ways. For example, if the contribution is a novel architecture, describing the architecture fully might suffice, or if the contribution is a specific model and empirical evaluation, it may be necessary to either make it possible for others to replicate the model with the same dataset, or provide access to the model. In general. releasing code and data is often one good way to accomplish this, but reproducibility can also be provided via detailed instructions for how to replicate the results, access to a hosted model (e.g., in the case of a large language model), releasing of a model checkpoint, or other means that are appropriate to the research performed.
        \item While NeurIPS does not require releasing code, the conference does require all submissions to provide some reasonable avenue for reproducibility, which may depend on the nature of the contribution. For example
        \begin{enumerate}
            \item If the contribution is primarily a new algorithm, the paper should make it clear how to reproduce that algorithm.
            \item If the contribution is primarily a new model architecture, the paper should describe the architecture clearly and fully.
            \item If the contribution is a new model (e.g., a large language model), then there should either be a way to access this model for reproducing the results or a way to reproduce the model (e.g., with an open-source dataset or instructions for how to construct the dataset).
            \item We recognize that reproducibility may be tricky in some cases, in which case authors are welcome to describe the particular way they provide for reproducibility. In the case of closed-source models, it may be that access to the model is limited in some way (e.g., to registered users), but it should be possible for other researchers to have some path to reproducing or verifying the results.
        \end{enumerate}
    \end{itemize}

\item {\bf Open access to data and code}
    \item[] Question: Does the paper provide open access to the data and code, with sufficient instructions to faithfully reproduce the main experimental results, as described in supplemental material?
    \item[] Answer: \answerYes{} 
    \item[] Justification: We will open-source the codebase, pre-trained weights, and the EgoTac-SC dataset upon publication. We have provided necessary material in the supplementary.
    \item[] Guidelines:
    \begin{itemize}
        \item The answer \answerNA{} means that paper does not include experiments requiring code.
        \item Please see the NeurIPS code and data submission guidelines (\url{https://neurips.cc/public/guides/CodeSubmissionPolicy}) for more details.
        \item While we encourage the release of code and data, we understand that this might not be possible, so \answerNo{} is an acceptable answer. Papers cannot be rejected simply for not including code, unless this is central to the contribution (e.g., for a new open-source benchmark).
        \item The instructions should contain the exact command and environment needed to run to reproduce the results. See the NeurIPS code and data submission guidelines (\url{https://neurips.cc/public/guides/CodeSubmissionPolicy}) for more details.
        \item The authors should provide instructions on data access and preparation, including how to access the raw data, preprocessed data, intermediate data, and generated data, etc.
        \item The authors should provide scripts to reproduce all experimental results for the new proposed method and baselines. If only a subset of experiments are reproducible, they should state which ones are omitted from the script and why.
        \item At submission time, to preserve anonymity, the authors should release anonymized versions (if applicable).
        \item Providing as much information as possible in supplemental material (appended to the paper) is recommended, but including URLs to data and code is permitted.
    \end{itemize}

\item {\bf Experimental setting/details}
    \item[] Question: Does the paper specify all the training and test details (e.g., data splits, hyperparameters, how they were chosen, type of optimizer) necessary to understand the results?
    \item[] Answer: \answerYes{} 
    \item[] Justification: Key implementation details are stated in Section 4.4, with comprehensive hyperparameters and data-split strategies provided in the appendix.
    \item[] Guidelines:
    \begin{itemize}
        \item The answer \answerNA{} means that the paper does not include experiments.
        \item The experimental setting should be presented in the core of the paper to a level of detail that is necessary to appreciate the results and make sense of them.
        \item The full details can be provided either with the code, in appendix, or as supplemental material.
    \end{itemize}

\item {\bf Experiment statistical significance}
    \item[] Question: Does the paper report error bars suitably and correctly defined or other appropriate information about the statistical significance of the experiments?
    \item[] Answer: \answerNo{} 
    \item[] Justification: The results are obtained from standard randomly seeded runs without cherrypicking, offering a fair and representative evaluation despite not including error bars.
    \item[] Guidelines:
    \begin{itemize}
        \item The answer \answerNA{} means that the paper does not include experiments.
        \item The authors should answer \answerYes{} if the results are accompanied by error bars, confidence intervals, or statistical significance tests, at least for the experiments that support the main claims of the paper.
        \item The factors of variability that the error bars are capturing should be clearly stated (for example, train/test split, initialization, random drawing of some parameter, or overall run with given experimental conditions).
        \item The method for calculating the error bars should be explained (closed form formula, call to a library function, bootstrap, etc.)
        \item The assumptions made should be given (e.g., Normally distributed errors).
        \item It should be clear whether the error bar is the standard deviation or the standard error of the mean.
        \item It is OK to report 1-sigma error bars, but one should state it. The authors should preferably report a 2-sigma error bar than state that they have a 96\% CI, if the hypothesis of Normality of errors is not verified.
        \item For asymmetric distributions, the authors should be careful not to show in tables or figures symmetric error bars that would yield results that are out of range (e.g., negative error rates).
        \item If error bars are reported in tables or plots, the authors should explain in the text how they were calculated and reference the corresponding figures or tables in the text.
    \end{itemize}

\item {\bf Experiments compute resources}
    \item[] Question: For each experiment, does the paper provide sufficient information on the computer resources (type of compute workers, memory, time of execution) needed to reproduce the experiments?
    \item[] Answer: \answerYes{} 
    \item[] Justification: We provide experimental details and computer resources both in the Section 4.4 and the appendix.
    \item[] Guidelines:
    \begin{itemize}
        \item The answer \answerNA{} means that the paper does not include experiments.
        \item The paper should indicate the type of compute workers CPU or GPU, internal cluster, or cloud provider, including relevant memory and storage.
        \item The paper should provide the amount of compute required for each of the individual experimental runs as well as estimate the total compute. 
        \item The paper should disclose whether the full research project required more compute than the experiments reported in the paper (e.g., preliminary or failed experiments that didn't make it into the paper). 
    \end{itemize}
    
\item {\bf Code of ethics}
    \item[] Question: Does the research conducted in the paper conform, in every respect, with the NeurIPS Code of Ethics \url{https://neurips.cc/public/EthicsGuidelines}?
    \item[] Answer: \answerYes{} 
    \item[] Justification: Our research stringently conforms to the NeurIPS Code of Ethics.
    \item[] Guidelines:
    \begin{itemize}
        \item The answer \answerNA{} means that the authors have not reviewed the NeurIPS Code of Ethics.
        \item If the authors answer \answerNo, they should explain the special circumstances that require a deviation from the Code of Ethics.
        \item The authors should make sure to preserve anonymity (e.g., if there is a special consideration due to laws or regulations in their jurisdiction).
    \end{itemize}

\item {\bf Broader impacts}
    \item[] Question: Does the paper discuss both potential positive societal impacts and negative societal impacts of the work performed?
    \item[] Answer: \answerYes{} 
    \item[] Justification: We have included a discussion regarding the positive impacts (e.g., embodied AI) and potential negative societal impacts in the supplementary material.
    \item[] Guidelines:
    \begin{itemize}
        \item The answer \answerNA{} means that there is no societal impact of the work performed.
        \item If the authors answer \answerNA{} or \answerNo, they should explain why their work has no societal impact or why the paper does not address societal impact.
        \item Examples of negative societal impacts include potential malicious or unintended uses (e.g., disinformation, generating fake profiles, surveillance), fairness considerations (e.g., deployment of technologies that could make decisions that unfairly impact specific groups), privacy considerations, and security considerations.
        \item The conference expects that many papers will be foundational research and not tied to particular applications, let alone deployments. However, if there is a direct path to any negative applications, the authors should point it out. For example, it is legitimate to point out that an improvement in the quality of generative models could be used to generate Deepfakes for disinformation. On the other hand, it is not needed to point out that a generic algorithm for optimizing neural networks could enable people to train models that generate Deepfakes faster.
        \item The authors should consider possible harms that could arise when the technology is being used as intended and functioning correctly, harms that could arise when the technology is being used as intended but gives incorrect results, and harms following from (intentional or unintentional) misuse of the technology.
        \item If there are negative societal impacts, the authors could also discuss possible mitigation strategies (e.g., gated release of models, providing defenses in addition to attacks, mechanisms for monitoring misuse, mechanisms to monitor how a system learns from feedback over time, improving the efficiency and accessibility of ML).
    \end{itemize}
    
\item {\bf Safeguards}
    \item[] Question: Does the paper describe safeguards that have been put in place for responsible release of data or models that have a high risk for misuse (e.g., pre-trained language models, image generators, or scraped datasets)?
    \item[] Answer: \answerNA{} 
    \item[] Justification: The introduced tactile prediction models and hand-object interaction datasets do not pose a high risk for malicious misuse.
    \item[] Guidelines:
    \begin{itemize}
        \item The answer \answerNA{} means that the paper poses no such risks.
        \item Released models that have a high risk for misuse or dual-use should be released with necessary safeguards to allow for controlled use of the model, for example by requiring that users adhere to usage guidelines or restrictions to access the model or implementing safety filters. 
        \item Datasets that have been scraped from the Internet could pose safety risks. The authors should describe how they avoided releasing unsafe images.
        \item We recognize that providing effective safeguards is challenging, and many papers do not require this, but we encourage authors to take this into account and make a best faith effort.
    \end{itemize}

\item {\bf Licenses for existing assets}
    \item[] Question: Are the creators or original owners of assets (e.g., code, data, models), used in the paper, properly credited and are the license and terms of use explicitly mentioned and properly respected?
    \item[] Answer: \answerYes{} 
    \item[] Justification: We properly cite all prior datasets used to construct our mixed corpus, and their respective licenses are respected and detailed in the supplementary material.
    \item[] Guidelines:
    \begin{itemize}
        \item The answer \answerNA{} means that the paper does not use existing assets.
        \item The authors should cite the original paper that produced the code package or dataset.
        \item The authors should state which version of the asset is used and, if possible, include a URL.
        \item The name of the license (e.g., CC-BY 4.0) should be included for each asset.
        \item For scraped data from a particular source (e.g., website), the copyright and terms of service of that source should be provided.
        \item If assets are released, the license, copyright information, and terms of use in the package should be provided. For popular datasets, \url{paperswithcode.com/datasets} has curated licenses for some datasets. Their licensing guide can help determine the license of a dataset.
        \item For existing datasets that are re-packaged, both the original license and the license of the derived asset (if it has changed) should be provided.
        \item If this information is not available online, the authors are encouraged to reach out to the asset's creators.
    \end{itemize}

\item {\bf New assets}
    \item[] Question: Are new assets introduced in the paper well documented and is the documentation provided alongside the assets?
    \item[] Answer: \answerYes{} 
    \item[] Justification: The newly introduced EgoTac-SC dataset is extensively documented in Section 3 and the appendix, and we will provide full documentation and scripts upon release.
    \item[] Guidelines:
    \begin{itemize}
        \item The answer \answerNA{} means that the paper does not release new assets.
        \item Researchers should communicate the details of the dataset\slash code\slash model as part of their submissions via structured templates. This includes details about training, license, limitations, etc. 
        \item The paper should discuss whether and how consent was obtained from people whose asset is used.
        \item At submission time, remember to anonymize your assets (if applicable). You can either create an anonymized URL or include an anonymized zip file.
    \end{itemize}

\item {\bf Crowdsourcing and research with human subjects}
    \item[] Question: For crowdsourcing experiments and research with human subjects, does the paper include the full text of instructions given to participants and screenshots, if applicable, as well as details about compensation (if any)? 
    \item[] Answer: \answerNA{} 
    \item[] Justification: The dataset collection was conducted locally without utilizing external crowdsourcing platforms.
    \item[] Guidelines:
    \begin{itemize}
        \item The answer \answerNA{} means that the paper does not involve crowdsourcing nor research with human subjects.
        \item Including this information in the supplemental material is fine, but if the main contribution of the paper involves human subjects, then as much detail as possible should be included in the main paper. 
        \item According to the NeurIPS Code of Ethics, workers involved in data collection, curation, or other labor should be paid at least the minimum wage in the country of the data collector. 
    \end{itemize}

\item {\bf Institutional review board (IRB) approvals or equivalent for research with human subjects}
    \item[] Question: Does the paper describe potential risks incurred by study participants, whether such risks were disclosed to the subjects, and whether Institutional Review Board (IRB) approvals (or an equivalent approval/review based on the requirements of your country or institution) were obtained?
    \item[] Answer: \answerNA{} 
    \item[] Justification: The local data collection strictly involved consenting volunteers performing mundane daily tasks, which does not pose risks requiring institutional review board (IRB) assessments beyond standard laboratory safety.
    \item[] Guidelines:
    \begin{itemize}
        \item The answer \answerNA{} means that the paper does not involve crowdsourcing nor research with human subjects.
        \item Depending on the country in which research is conducted, IRB approval (or equivalent) may be required for any human subjects research. If you obtained IRB approval, you should clearly state this in the paper. 
        \item We recognize that the procedures for this may vary significantly between institutions and locations, and we expect authors to adhere to the NeurIPS Code of Ethics and the guidelines for their institution. 
        \item For initial submissions, do not include any information that would break anonymity (if applicable), such as the institution conducting the review.
    \end{itemize}

\item {\bf Declaration of LLM usage}
    \item[] Question: Does the paper describe the usage of LLMs if it is an important, original, or non-standard component of the core methods in this research? Note that if the LLM is used only for writing, editing, or formatting purposes and does \emph{not} impact the core methodology, scientific rigor, or originality of the research, declaration is not required.
    \item[] Answer: \answerNA{} 
    \item[] Justification: We did not use LLMs as a core or original method component in this research.
    \item[] Guidelines:
    \begin{itemize}
        \item The answer \answerNA{} means that the core method development in this research does not involve LLMs as any important, original, or non-standard components.
        \item Please refer to our LLM policy in the NeurIPS handbook for what should or should not be described.
    \end{itemize}

\end{enumerate}

\end{document}